\documentclass[journal,compsoc]{IEEEtran} 
\usepackage{graphicx}
\graphicspath{{./images/}}
\usepackage{subfigure}
\usepackage{stackengine}
\usepackage{multirow}
\usepackage{amsmath,amsfonts,amssymb,amsthm}
\usepackage{url}
\usepackage{floatrow}
\usepackage{float}
\usepackage{enumerate}
\usepackage{wrapfig}
\usepackage{float}
\usepackage{balance}
\usepackage[percent]{overpic}
\usepackage{graphicx}
\usepackage[ruled]{algorithm2e}
\usepackage{xcolor}
\usepackage{mathbbol}
\usepackage{floatrow}
\usepackage{wrapfig}
\usepackage{tikz,pgfplots}
\usepackage{rotating}
\usepackage{import}
\newfloatcommand{capbtabbox}{table}[][\FBwidth]

\newfloatcommand{capbfigbox}{figure}[][\FBwidth]

\newcommand{\ie}{\emph{i.e.}}

\newcommand{\eg}{\emph{e.g.}}

\newcommand{\tr}{{\rm tr}}
\newcommand{\W}{\mathbf{W}}
\newcommand{\kk}{{(\kappa)}}
\newcommand{\X}{\mathbf{X}}
\newcommand{\Y}{\mathbf{Y}}

\newcommand{\f}{\mathbf{f}}
\newcommand{\pphi}{\pmb{\phi}}
\newcommand{\kea}{\varphi_{\sf{kron}-{\displaystyle e} }}
\newcommand{\kpa}{\varphi_{\sf{kron}-{\displaystyle \pi}}}
\newcommand{\var}{\mathbb{var}}
\newcommand{\pphiP}{\pmb{\phi}_{\sf{kron}-{\displaystyle \pi}}}
\newcommand{\pphiE}{\pmb{\phi}_{\sf{kron}-{\displaystyle e}}}

\newtheorem{thm}{Theorem}
\newtheorem{defn}{Definition}

\newtheorem{cor}{Corollary}
\ifCLASSINFOpdf
\else
\fi

\ifCLASSOPTIONcompsoc
\usepackage[nocompress]{cite}
\else
\usepackage{cite}
\fi

\begin{document}
%
\title{Scalable and Compact 3D Action Recognition with Approximated RBF Kernel Machines}
%
%
%

\author{Jacopo Cavazza, Pietro Morerio, and Vittorio 
Murino,~\IEEEmembership{Senior Member,~IEEE}

\thanks{\scriptsize J. Cavazza, P. Morerio and V. Murino are with Pattern Analysis \& 
Computer Vision (PAVIS), Istituto Italiano di Tecnologia (IIT), Genova, Italy.}
\thanks{\scriptsize J. Cavazza is also with Dipartimento di Ingegneria Navale, 
Elettrica, Elettronica e delle Telecomunicazioni (DITEN), University of Genoa, Italy.}
\thanks{\scriptsize V. Murino is also with Dept. of Computer Science, University of 
Verona, Italy.}
\thanks{Primary email contact: \texttt{jacopo.cavazza@iit.it}.}
}

%
%

\markboth{IEEE Transactions on Pattern Analysis and Machine Intelligence}%
{Jacopo Cavazza \MakeLowercase{\textit{et al.}}:  Approx. Kernel for Action 
Recognition}
%






%

\IEEEtitleabstractindextext{\begin{abstract}
Despite the recent deep learning (DL) revolution, kernel machines still remain powerful methods for action recognition. DL has brought the use of large datasets and this is typically a problem for kernel approaches, which are not scaling up efficiently due to kernel Gram matrices. Nevertheless, kernel methods are still attractive and more generally applicable since they can equally manage different sizes of the datasets, also in cases where DL techniques show some limitations. This work investigates these issues by proposing an explicit approximated representation that, together with a linear model, is an equivalent, yet scalable, implementation of a kernel machine. Our approximation is directly inspired by the exact feature map that is induced by an RBF Gaussian kernel but, unlike the latter, it is finite dimensional and very compact. We justify the soundness of our idea with a theoretical analysis which proves the unbiasedness of the approximation, and provides a vanishing bound for its variance, which is shown to decrease much rapidly than in alternative methods in the literature. 
In a broad experimental validation, we assess the superiority of our approximation in terms of 1) ease and speed of training, 2) compactness of the model, and 3) improvements with respect to the state-of-the-art performance. 
	\end{abstract}	

	}



%
\maketitle

\IEEEraisesectionheading{\section{Introduction}}

\IEEEPARstart{A}{ction} recognition is a paramount research domain in machine intelligence and computer vision, being nowadays ubiquitous in many application domains such as human-robot interaction, autonomous driving, elderly care and video-surveillance, just to name a few \cite{survey}. Yet, major difficulties arise when dealing with videos due to general visual ambiguities such as illumination variations, the presence of clutter/noise in the scene, occlusions or unfavorable recording viewpoint. 
Moreover, the variability of action evolution, as either executed by different human subjects or implicit in the structure of the action execution, further contributes to complicate the classification process. Fortunately, the adoption of novel range sensors constitutes an effective countermeasure as they provide alternative data to process, more robust to the above mentioned issues. 
Actually, this type of sensors (e.g. Kinect) also allows to represent a given action -- other than by dense range data -- as a collection of skeletal joint positions progressing in time, through real-time algorithms \cite{shotton2016human}.
Action recognition can thus be reformulated as the problem of classifying the multivariate time-series $\mathbf{P} \in \mathbb{R}^{3J \times T}$, which collect the three-dimensional coordinates of the $J$ skeletal joints positions over $T$ temporal acquisitions. 

Within the data structure $\mathbf{P}$, $J$ is fixed by the selection of the device which acquires the joints (\eg, Kinect or VICON), while $T$ typically changes across instances. Therefore, a minimal requirement for encoding this data is to be invariant  to the variability of $T$. Among the possible feature encoding methods (see \cite{survey} for a literature review), the symmetric and positive definite (SPD) covariance (COV) operator guarantees this property, while also demonstrated to score a solid performance in 3D action recognition \cite{Wang:ICCV15,Harandi:CVPR14,Cavazza:ICPR16 ,ker_approx,kermeetfeat}. Actually, in addition to properly modeling the skeletal dynamics with a second order statistics, the COV operator is also naturally able to handle different temporal durations of the action instances. 
This avoids slow pre-processing stages such as time warping or interpolation \cite{Vemulapalli:CVPR14}, needed to ``re-align'' the different sequences before the actual classification. Moreover, performance achieved by COV-based methods are always comparable and sometimes superior to the one achieved by deep learning methods \cite{Shahroudy:CVPR16,Liu:ECCV16,Liu:17,Ke:CVPR17,SPDnet,deeplie,JCNN1,JCNN2}, 
which, instead, typically require a massive amount of data and large computational power (on GPUs) for training.

All covariance-based paradigms for action recognition can be framed as the problem of classifying $d \times d$ data instances $\X$. In the case of skeleton data, $d = 3J$ and $\X = \tfrac{1}{T-1} \mathbf{P} \mathbf{J} \mathbf{P}^\top$, where $\mathbf{J} = \tfrac{1}{T} \mathbf{I} - \boldsymbol{1}_{T \times T}$ (being \textbf{I} the identity matrix) is the centering matrix as defined in \cite{Minh:NIPS14,Minh:CVPR16}. To accomplish such task, kernel theory \cite{Scholkopf:02} naturally promotes max-margin approaches in order to learn decision boundaries maximally separating (action) classes. Interestingly, this can be done by \emph{only} evaluating a kernel function $K$ that, in our work, is fixed as the Radial Basis Function (RBF) Gaussian kernel:
\begin{equation}\label{eq:K}
K(\mathbf{X},\mathbf{Y}) = \exp \left( - \dfrac{1}{2 \sigma^2} \|  
\mathbf{X} - \mathbf{Y} \|_F^2\right).
\end{equation}
The choice of this kernel is motivated by a set of beneficial properties, \ie, 1) invariance to translations, 2) isotropy and 3) infinite-smoothness. Moreover, due to its robustness with respect to the parameter $\sigma$, it has been broadly and effectively used in the literature for many tasks
\cite{Scholkopf:02,RR:NIPS07,KK:AISTATS13,Vedaldi:BMVC,Vedaldi_tPAMI,Minh:NIPS14,Ring:PRL16,Minh:CVPR16}. 
More specifically, when applying the change of variables $\X = \log(\tfrac{1}{T-1} \mathbf{P} \mathbf{J} \mathbf{P}^\top )$, equation \eqref{eq:K} becomes the log-Euclidean kernel, which, thanks to its strong theoretical properties, is well suited to compare SPD matrices \cite{Arsigny:Siam:07}. To this end, it has been widely exploited in computer vision and related fields, such as action recognition \cite{Cavazza:ICPR16} or pedestrian re-identification \cite{Tosato}, to name a few. 

Unfortunately, this approach has a limited scalability, since \eqref{eq:K} has to be computed for each pair of examples within the training set  $\{\X_1,\dots,\X_N\}$ and for each ordered 
pair across training and test sets $\{\Y_1,\dots,\Y_M\}$. This yields to the training and test Gram matrices $K(\X_i,\X_j)$ and $K(\Y_k,\X_i)$, $i,j = 1,\dots,N$ and $k = 1,\dots,M$, respectively. 
In the case of large number of samples $M$ and/or $N$, Gram matrices are quite hard to both store and manipulate when performing the optimization to determine the decision boundaries. For instance, if $M,N \sim 10^4$, about $10^{12}$ products are required to perform a matrix inversion, which will likely result in an out-of-memory error. 

Such problem can be circumvented if we are able to obtain an explicitly computable feature representation $\pphi$ such that $\langle \pphi(\X), \pphi(\Y) \rangle$ equals \eqref{eq:K}, even approximately. In fact, while a linear machine fed with $\pphi$ is theoretically equivalent to a kernel machine (thanks to the kernel trick \cite{Scholkopf:02}), training 
a linear SVM is scalable even in the big data regime, differently from an exact kernel SVM \cite{liblinear,liblinearGPU1,liblinearGPU2}. However, despite a few approximation schemes have been proposed \cite{RR:NIPS07,KK:AISTATS13,Vedaldi:BMVC,Vedaldi_tPAMI,Fastfood}, there is not yet a definitive answer about which performs the best in applicative settings.
  
With respect to all the problems presented above, our paper introduces the following contributions.

\begin{enumerate}
	\item We propose a novel, explicit \emph{random} feature map, which can
	rigorously be interpreted as a compact approximation inspired by the exact (and 
	infinite-dimensional) feature encoding induced by \eqref{eq:K}.
	\item We theoretically show that, marginalizing the sources of randomness, the proposed estimator of \eqref{eq:K} is unbiased, and its variance has an explicit upper bound that is i) more clearly interpretable and ii) more rapidly decreasing as a function of the size of the approximation. These properties make our approach more favorable with respect to  competing methods in the literature \cite{RR:NIPS07,KK:AISTATS13,Vedaldi:BMVC,Vedaldi_tPAMI,Fastfood}.
	\item We present an extensive comparative analysis between existing \textit{approximation schemes} and our proposed approach, proving its superiority. 
    \item In a broad experimental validation on a consistent number of publicly available benchmark datasets, we demonstrate the advantages of our method for 3D action recognition, in terms of ease and speed of training, compactness of the representation and \textit{improvement over state-of-the-art performance}.
\end{enumerate}

The rest of the paper is structured as follows. Section \ref{sez:RW} recaps the relevant works in action recognition and kernel approximation. In Section \ref{sez:kron-E}, we dissect the proposed approximation in formal terms. The experimental validation is presented in Sections \ref{sez:exp} and \ref{sez:soa}. 
Finally, Section \ref{sez:conc} draws conclusions, profiles limitations and sketches the future work.

\section{Related Work}\label{sez:RW}

In this Section, we discuss some of the most relevant related works in the field of 3D human action recognition, focusing on state-of-the art approaches in (approximated) kernel methods and feature learning. 

{\bf Kernel methods.} Within 3D action recognition methods on manifolds, a major role is played by symmetric and positive definite (SPD) matrices and, among them, covariance operators. 
The latter are either extended to the infinite dimensional case \cite{Harandi:CVPR14} or hierarchically combined in a temporal pyramid \cite{egizi}. The conceptual analogy with trial-specific kernel matrices is investigated \cite{Wang:ICCV15,ECCV16}, whereas kernelized covariance allow to model arbitrary non-linear relationships \cite{Cavazza:ICPR16}.

Alternatively, Hankel matrices proficiently model action dynamics. They are used in tandem with a Hidden Markov Model \cite{Camps:ACCV14} or a Riemannian nearest neighbors with class-prototypes \cite{Camps:CVPR16}. As a slightly different paradigm, the Lie group \cite{Vemulapalli:CVPR14} and associated Lie algebra \cite{Vemulapalli:CVPR16} of the special Euclidean group of roto-translations are very effective in classifying skeletal joints temporal sequences.

However, as already mentioned, kernel methods usually do not scale up easily to big datasets due to demanding storage and computational costs. As a solution, the exact kernel representation can be replaced by an approximated, more efficient one. In the literature, this is done essentially in two ways\footnote{For a broad review of approximated feature map for kernel machine, please refer to \cite[\S 4]{Ring:PRL16}.}, as described in the following.\\
$i)$ The kernel Gram matrix is replaced with a surrogate low-rank version. Within these methods, \cite{Bach:ICML05} applied Cholesky decomposition and \cite{Zhang:ICML08} adopted Nystr\"{o}m approximation. \\
$ii)$ Instead of the exact kernel function $k$, an explicit feature map $\phi$ is computed, so that the induced linear kernel $\langle \phi(\mathbf{x}),\phi(\mathbf{y}) \rangle$ approximates $k(\mathbf{x},\mathbf{y})$. Our work belongs to this class of approaches. 
\\
Within the latter proposed methods, Rahimi \& Recht \cite{RR:NIPS07} and later Vedaldi et al. \cite{Vedaldi:BMVC,Vedaldi_tPAMI}, exploit the formalism of the Fourier Transform to approximate shift invariant kernels $k(\mathbf{x},\mathbf{y}) = k(\mathbf{x}-\mathbf{y})$ through an expansion of trigonometric functions. 
Leveraging on a similar idea, Le et al. \cite{Fastfood} speed up the computation by exploiting the Walsh-Hadamard transform, downgrading the running cost of \cite{RR:NIPS07} from linear to log-linear with respect to the data dimension. 
Recently, Kar \& Karnick \cite{KK:AISTATS13} have proposed an approximated feature maps for dot product kernels $k(\mathbf{x},\mathbf{y}) = k(\langle \mathbf{x},\mathbf{y} \rangle)$ by leveraging on the Taylor expansion of $k$.
 
{\bf Feature learning.} The representation for skeletal joints can be learned from the data itself. Du et al. \cite{Du:CVPR15} propose a hierarchy of bidirectional recurrent neural networks to represent in a bottom-up fashion all the structural relationships between joints in the human skeleton. Starting from legs, arms and torso, modeled with separated networks, higher levels of the hierarchy aggregate all parts while a final softmax layer is responsible for the final action classification. Long-Short Term Memory (LSTM) models can be proficiently applied to 3D action recognition. 
Indeed, after the introduction of the first modern big-size dataset for 3D action recognition from joints \cite{Shahroudy:CVPR16}, the performance of LSTM networks achieves the state-of-the-art level by either performing a direct training on the raw joint coordinates of the human body \cite{Shahroudy:CVPR16} or implementing the true human skeleton structure with a direct acyclic graph \cite{Liu:ECCV16} and, eventually, recurring to attention mechanisms \cite{Liu:CVPR17}. Recently, multiple deep RNN \cite{RNNtree:ICCV17} and LSTM \cite{TSLSTM:ICCV17} have been combined in an adaptive tree-structure for hierarchical classification.
Alternatively, joint trajectories are used to produce distance maps, then converted into images to fine-tune convolutional neural networks (CNN), which can be therefore applied for 3D action recognition \cite{JCNN1,JCNN2,Ke:CVPR17}.

\subsection{Originality aspects}

In this work, we frame covariance-based action recognition into the general problem of kernel machine training. 
Since the latter suffers from scalability issues, we propose an explicit feature encoding that, combined with a linear model, is able to implement an exact kernel machine. 
With respect to the analogous approaches in the literature \cite{RR:NIPS07,Vedaldi:BMVC,Vedaldi_tPAMI,Fastfood,KK:AISTATS13}, our method is corroborated by a stronger theoretical guarantee. Specifically, with respect to state-of-the-art kernel methods for action recognition \cite{Wang:ICCV15,ECCV16,Vemulapalli:CVPR14,Cavazza:ICPR16,Vemulapalli:CVPR16}, our model is more compact and scalable. When compared with deep learning methods \cite{Shahroudy:CVPR16,Liu:ECCV16,JCNN1,JCNN2,Ke:CVPR17,Liu:CVPR17}, our pipeline is easier and faster to train, and reaches comparable performance.

This approach extends two previous works \cite{ker_approx,kermeetfeat}. With respect to \cite{ker_approx}, we generalize the class of feature maps thereby proposed and show that \cite{ker_approx} is a particular case of our approximation. Besides, we extend its experimental validation on a new dataset (the NTU RGB+D) and discuss its computational cost. 
With respect to \cite{kermeetfeat}, we show that the architecture thereby proposed can be framed into our theoretical analysis, as a particular case. This provides an insight on the reason why \cite{kermeetfeat} achieves solid performance. Further, we extend its experimental validation on new datasets with state-of-the-art comparisons.

\section{Approximating the RBF kernel with Kronecker products}\label{sez:kron-E}

In this Section, we present in formal terms our original technique to approximate the RBF 
kernel \eqref{eq:K} by means of a low-dimensional and explicit feature map, characterized by a random component which is ultimately responsible of the quality of the approximation itself. Indeed, when averaging upon all the possible realization of such component, our representation approximates \eqref{eq:K} with zero bias. Additionally, the variance of such estimation can be controlled by an explicit upper bound that easily writes as a function which rapidly decreases as the feature dimensionality increases.

\subsection{Construction of the approximated feature map} Given $\X \in 
\mathbb{R}^{d \times d}$ and fixed a strictly positive integer $\nu$, that 
corresponds to the feature dimensionality, our approximation is defined as follows.

\begin{defn}
We define a $\nu$ dimensional vector 
$\pphi_{\sf{kron}-{\displaystyle \pi} }(\X)$ whose components $\phi_{\sf{kron}-{\displaystyle \pi},1}(\X), \dots, 
\phi_{\sf{kron}-{\displaystyle \pi},\nu}(\X)$ are ($1/\sqrt{\nu}$-multiplied) independent realizations of the following scalar function 
\begin{equation}\label{eq:varphi_kron-P}
\varphi_{\sf{kron}-{\displaystyle \pi}}(\X) =  \dfrac{1}{\sigma^{2n}} \sqrt{\dfrac{ 
		\exp(-\tfrac{1}{\sigma^2})}{
		\rho(n) n!}} \tr\left( \otimes_{\kappa = 1}^n 
{\W^\kk}^\top \X \right).
\end{equation}
In \eqref{eq:varphi_kron-P}, $\sigma > 0$ defines the bandwidth of the kernel function \eqref{eq:K}, $n$ is sampled from \emph{any} distribution $\rho$ supported over the integers. Furthermore, the following 
assumptions are made:
\begin{enumerate}[{A}.1]
	\item ${\W^\kk}$ are (elementwise) drawn from the distribution $\mathcal{P}$ 
	with null expected value and standard deviation equals to the kernel's 
	bandwidth $\sigma$.
	\item The $d \times d$ matrix which is inputted to $\kpa$ lies on the Frobenius 
	norm-unitary sphere, that 
	is $\| \X \|_F = 1.$ 
\end{enumerate}
\end{defn}
Note that the $\kpa(\mathbf{X})$ has two sources of randomness. First, the integer $n$, which is sampled from $\rho$. Second, precisely $n$ matrices $\W^{(1)},\dots,\W^\kk,\dots,\W^{(n)}$ are sampled, so that each of their element is independently drawn from $\mathcal{P}$. More in detail, for each $\kappa = 1,\dots,n$, the transpose of $\W^\kk$ is (row-by-column) multiplied by $\X$. Afterwards, the results of the previous operation are combined together with a Kronecker product and, finally, the trace operator is evaluated. For the sake of clarity, let us notice that, since the trace operator applied on matrix returns a scalar, $\kpa(\X) \in \mathbb{R}$ and $\pphi_{\sf{kron}-{\displaystyle \pi} }(\X) \in \mathbb{R}^\nu$, since it stacks $\nu$ independent  realizations of $\kpa(\X)$ (divided by $\sqrt{\nu}$, which is factorized out of the definition of $\varphi$ only for convenience in the demonstrations). Algorithm \ref{alg:kronP} provides the pseudo-code for the construction process.

With respect to the assumptions A.1 and A.2, the first one constrains the distribution $\mathcal{P}$. Indeed, let us notice that, in all our theoretical exposition, the distributions $\rho$ and $\mathbf{P}$ are allowed to be highly general, and we will specify them only in the experiments when we need to numerically sample from them. For instance, A.1 is satisfied if $\mathcal{P} = \mathcal{N}(0,\sigma^2)$, being fixed as a zero-mean Gaussian with $\sigma^2$ variance. 

Instead, A.2 is only technical and does not really represent a constraint under an applicative point of view. Indeed, given an arbitrary input data $\X$, we can achieve A.2 by dividing $\X$ entrywise by $\| \X \|_F$. Such operation is easy to perform and it is along the line of the classical pre-processing which is applied on the data before passing them to a kernel method - as for instance, the component-wise division by the standard deviation is a common preprocessing step before SVM training \cite{Scholkopf:02}. If compared with similar results in \cite{RR:NIPS07,Vedaldi:BMVC,Vedaldi_tPAMI,KK:AISTATS13,Fastfood}, the assumption of unitary norm for $\X$ and $\Y$ is in line with the analogous assumptions of sampling the data from a given submanifold - with the remarkable difference that our assumption is easy to satisfy also in an applicative domain.

\begin{algorithm}[t!]
	\KwIn{A normalized $d \times d$ input matrix $\mathbf{X}$, the desired feature size $\nu$, the 
		probability distributions $\rho$ over integers and $\mathcal{P}$ over real numbers, the kernel bandwith $\sigma > 0$.}
	\KwOut{$
		[\phi_{\sf{kron}-{\displaystyle \pi},1}(\X),\dots,\phi_{\sf{kron}-{\displaystyle \pi},\nu}(\X)]$} 
	\ForEach{$j = 1,\dots,\nu$}{
		\nl Sample $n$ according to $\rho$\\
		\ForEach{$\kappa = 1,\dots,n$}{
			\nl Sample $\W^\kk \in \mathbb{R}^{d \times d}$ from $\mathcal{P}$ 
			elementwise.
		}
		\nl Compute the scalar $\pi(\X) = \tr\left( \otimes_{\kappa = 1}^n 
		{\W^\kk}^\top \X \right)$ \\
		\nl {\bf Return} $\phi_{\sf{kron}-{\displaystyle \pi},j}(\X) = \sigma^{-2n} 
		\left( \tfrac{\exp(-\sigma^{-2})}{\nu \rho(n) n!} \right)^{1/2} 
		\pi(\X)$}
	\caption{Approx, by Kronecker product.}
	\label{alg:kronP}
		\vspace{.3 cm}
\end{algorithm}

Before digging into the details of the theoretical foundation, lets us provide the intuition behind equation \eqref{eq:varphi_kron-P}.

\vspace{4 pt}

\subsection{Intuition behind the genesis of  $\varphi_{\sf{kron}-{\displaystyle \pi}}$}
\label{sez:intuition} According to the well established kernel theory 
\cite{Scholkopf:02}, the exact feature map $\f$ associated to the RBF kernel \eqref{eq:K} is infinite-dimensional. Still, it can be expressed in closed form. In fact, without loss of generality, let us assume $d = 1$ and, for the sake of simplicity, let $\sigma = 1$. Consequently, we replace the matrices $\X,\Y$ with the scalars $x,y$ and, in such a case, the kernel function \eqref{eq:K} rewrites as $K(x,y) = \exp( - \tfrac{1}{2} (x - y)^2)$.

We would like to write the exact infinite dimensional feature map $x \mapsto \f(x)$ for such  RBF kernel, i.e. the exact infinite-dimensional vector $\f(\cdot)$ such that
\begin{equation}
\langle \f(x),\f(y) \rangle = K(x,y) = \exp( -(x - y)^2/2) 
\end{equation}
where the inner product $\langle \cdot, \cdot \rangle$ is computed over the square-integrable series of $\f(\cdot)$.
Since 
\begin{equation}
\exp( -(x - y)^2/2) = \exp(-x^2/2) \cdot \exp(xy) \cdot \exp(-y^2/2),
\end{equation}
we can take advantage of the Taylor expansion to obtain
\begin{equation}\label{eq:fexact}
\f(x) = \sqrt{e^{-x^2}}\left[   1, 
x, \dfrac{x^2}{\sqrt{2!}}, \dfrac{x^3}{\sqrt{3!}}, \dots, \dfrac{x^n}{\sqrt{n!}}, \dots \right].
\end{equation}

As certified by Lagrange's remainder formula for Taylor expansions \cite{Rudin}, a good approximation of \eqref{eq:fexact} is obtained by considering all the terms which are less or equal to an arbitrary degree $n$. In the scalar case, these terms are exactly $n$. 
Differently, in order to compute the products for $d >1$, the terms of a given degree $n$ must include all the possible combinations $X_{11}^{\alpha_{11}} X_{12}^{\alpha_{12}} \cdots X^{\alpha_{ij}}_{ij} \cdots X_{dd}^{\alpha_{dd}}$, where $X_{ij}$ are the components of $\X$ and $\alpha_{ij}$ are $d^2$ non-negative integers such that $\sum_{ij} \alpha_{ij} = n$. That is, we have to consider all the $n / \prod_{ij}\alpha_{ij}!$ combinations, and this has an exponential complexity with respect to $d$ (check \cite[page 39.]{noteML}). This clearly produces an exponentially-sized feature map that, as shown in \cite{Ring:PRL16}, is formally fine but obviously not applicable in real-world datasets. In fact, as operative condition assumed in \cite{Ring:PRL16}, $d$ needs to be less than 4.

Since the analytical pipeline inspired by Taylor's remainder theorem is not viable in practical pattern analysis, in this work we propose a manageable alternative solution. When asked to build a $\nu$-dimensional representation, we repeat $\nu$ times the following pipeline. We sample $n$ from $\rho$ and we use $n$ as a pointer to index which component of \eqref{eq:fexact} to sample. Then, as a surrogate technique for computing all the possible combinations of products of degree $n$, we introduce $ \otimes_{\kappa = 1}^n {\W^\kk}^\top \X = \W^{(1)\top} \otimes \X \otimes \dots \W^{(n)\top} \otimes \X$. The latter is directly inspired from the technique of random rescaling, which is common practice in random approximated feature map approaches \cite{RR:NIPS07,Vedaldi:BMVC,Vedaldi_tPAMI,Fastfood,KK:AISTATS13}, where introducing random projections can be interpreted as a trick to "recover" from the sparse sampling of $n$. In the limit case where $\W^\kk$  are identity matrices, $ \otimes_{\kappa = 1}^n {\W^\kk}^\top \X = \X^{\otimes n}$ and we can find a clear analogy between scalar exponentiation in \eqref{eq:fexact} and Kronecker exponentiation in \eqref{eq:varphi_kron-P}, being the latter a $d \times d$ generalization of the former.

\subsection{Unbiasedness and variance bound}\label{sez:proofs}

In this Section, we demonstrate that, thanks to assumptions A.1 and A.2, once averaging upon all possible realizations of $n$ from $\rho$ and $\W^{(1)},\dots,\W^{(n)}$ from $\mathcal{P}$, we have no bias in approximating the kernel - that is, the expected value of our $\langle \pphiP(\X),\pphiP(\Y) \rangle$ coincides with \eqref{eq:K} and, at the same time, we are able to control the variance of the estimation.

{\bf Unbiasedness of  $\pphi_{\sf{kron}-{\displaystyle \pi} }$.} As previously explained, an exact feature map $\f$ is able to satisfy the equality $\langle \f(\X), \f(\Y) \rangle = K(\X,\Y)$. Thanks to the well established kernel trick \cite{Scholkopf:02}, one does not need to compute $\f$ explicitly but, instead, a kernel machine can be trained by evaluating the kernel function only. In many cases (like the one of RBF kernel \eqref{eq:K}), computing $\f$ explicitly is impossible due to its infinite dimension. Moreover, on the opposite, computing the kernel function does not scale to big datasets, since evaluating $K(\X,\Y)$ for \emph{every} $\X$ and $\Y$ has a quadratic complexity. Due to the prohibitive size of the Gram matrices, either the training or inference stages may be simply not computationally affordable (typically because of out-of-memory issues).

In order to accommodate for that, we propose to replace $\f$ with a map $\pphi_{\sf{kron}-{\displaystyle \pi} }$, such that 
\begin{equation}\label{approx}
\langle \pphi_{\sf{kron}-{\displaystyle \pi} }(\X), \pphi_{\sf{kron}-{\displaystyle \pi} }(\Y) \rangle \approx K(\X,\Y)
\end{equation}
with the crucial difference that $\pphi$ is explicitly computable. In other words, while the kernel trick allows to replace the feature map $\f$ with the kernel function $K$, we revert the perspective, and evaluate the kernel function with  $\pphi_{\sf{kron}-{\displaystyle \pi} }$, which, differently from $\f(\X)$, is finite-dimensional and explicitly computable. In fact, a linear model fed with $\pphiP$ is a theoretically valid estimate for the exact kernel machine fed with \eqref{eq:K}. 


As well established in the literature that similarly proposed random approximated feature maps \cite{RR:NIPS07,KK:AISTATS13,Fastfood,Vedaldi:BMVC,Vedaldi_tPAMI}, we want to demonstrate the validity of the approximation by showing that, once averaging upon all the sources of randomness which affect our feature map $\pphi_{\sf{kron}-{\displaystyle \pi} }$, an equality holds in eq. \ref{approx}. In other words, we want to prove the absence of biases in the approximation.

\begin{thm}[Unbiased approximation for 
	$\pphi_{\sf{kron}-{\displaystyle \pi}}$]\label{thm:kronP-Exp}
	With the previous notations, the linear kernel $\langle 
	\pphi_{\sf{kron}-{\displaystyle \pi}}(\X),\pphi_{\sf{kron}-{\displaystyle \pi}}(\Y)\rangle$ 
	induced by $\kpa$ is an unbiased estimator for $K(\X,\Y)$ as in 
	\eqref{eq:K}. Indeed,
	\begin{equation}
	\mathbb{E}_{n,\mathcal{P}} \left[ \langle 
	\pphi_{\sf{kron}-{\displaystyle \pi}}(\X),\pphi_{\sf{kron}-{\displaystyle \pi}}(\Y)\rangle \right] = 
	K(\X,\Y),
	\end{equation}
	being the expected value jointly computed over all possible realizations of $n$ from 
	$\rho$ and of $\W^\kk$ from $\mathcal{P}$, $\kappa = 1,\dots,n$.
\end{thm}

\proof See the Supplementary Material, Section 1. \endproof

\vspace{4 pt}

{\bf Bound on the variance for  $\pphi_{\sf{kron}-{\displaystyle \pi}}$.} Theorem \ref{thm:kronP-Exp} guarantees that, on average, $\langle \pphi_{\sf{kron}-{\displaystyle \pi}}(\X), \pphi_{\sf{kron}-{\displaystyle \pi}}(\Y) \rangle$ is a good approximation for $K(\X,\Y)$, since there is no bias. This is a strong and necessary assumption to ensure that our statistical estimator is reliable, but it does not take into account the variance, i.e. the \textit{quality} of the approximation. Namely, even an unbiased estimator can heavily deviate from its expected value if there are no theoretical guarantees for its variance.
We can prove that our estimator well behaves also in this respect, since $\pphi_{\sf{kron}-{\displaystyle \pi}}$ induces a linear kernel whose variance can be upper bounded as follows.

\begin{thm}[Bound on the variance of $\pphi_{\sf{kron}-{\displaystyle \pi}}$]\label{thm:kronP-Var}
	With the previous notation, the linear kernel $\langle \pphi_{\sf{kron}-{\displaystyle \pi}}(\X), 
	\pphi_{\sf{kron}-{\displaystyle \pi}}(\Y)\rangle$ induced by $\kpa$ has a controlled variance 
	which 
	is bounded by a linear function of the feature dimensionality $\nu$. Precisely,
	\begin{equation}\nonumber
	\var\left[ \langle \pphi_{\sf{kron}-{\displaystyle \pi}}(\X), 
	\pphi_{\sf{kron}-{\displaystyle \pi}}(\Y)\rangle \right] \leq 
	\dfrac{C_\rho}{\nu^3} 
	\exp\left(\dfrac{9 \, m_4(\mathcal{P}) - 2\sigma^4}{\sigma^8}\right)
	\end{equation}
	where the variance is computed over all possible realizations of $n$ from $\rho$ and all possible manners of sampling $\W^\kk$ from $\mathcal{P}$ for each $\kappa$. 
    
    $C_\rho$ is defined as 
	\begin{equation}\label{eq:Crho}
	C_\rho = \sum_{n = 0}^{\infty} \dfrac{1}{\rho(n) \cdot n!}
	\end{equation}
	 and $m_4(\mathcal{P})$ 
	denotes the fourth order moment of $\mathcal{P}$.
\end{thm}

\proof See the Supplementary Material, Section 2. \endproof

If we neglect the function $\exp\left(\tfrac{9 \, m_4(\mathcal{P}) - 2\sigma^4}{\sigma^8}\right)$, which is fixed after we select $\mathcal{P}$ and the bandwidth $\sigma$ in \eqref{eq:K}, the boundary on the variance rewrites as $C_\rho / \nu^3$. This means that, as the feature dimension $\nu$ increases, the variance very sharply converges to zero as $1/\nu^3$, i.e. our approximation converges to its expected value.

The constant $C_\rho$ may however affect the quality of this limit. For instance, if we choose $\rho$ to be a Geometric distribution of parameter $0 < \theta \leq 1$, we have $\rho(n) = (1 - \theta)^n \theta$ and one can analytically obtain
\begin{equation}
C_\rho = \dfrac{1 - \theta}{\theta} \exp\left(\dfrac{1 - \theta}{\theta}\right).
\end{equation}
The previous function increases and diverges for $\theta \to 1^-$ and $\theta \to 0^+$ making the bound potentially loose. The limit case $\theta \approx 0$ is very unfavorable also in in practice: in such a case a value sampled from $\rho$ is high with high probability and, therefore, many Kronecker products need to be evaluated in \eqref{eq:varphi_kron-P}. On the opposite side, the case $\theta \approx 1$ is very favorable in practical terms since $n$ is small with high probability and therefore the cost of computing \eqref{eq:varphi_kron-P} approaches the minimal one. Further considerations on the practical choice of $\theta$ are also reported in Section \ref{sez:cost}. 

To conclude our discussion on the variance, we provide the following result, which is derived from Theorem \ref{thm:kronP-Exp} and \ref{thm:kronP-Var} as a straightforward consequence of Chebyshev inequality. 

\begin{cor}\label{cor:kronP}
Under the previous hypothesis, for any $\epsilon > 0$ and $\X,\Y$ $d \times d$ matrices, the probability $\mathbb{P} \left[ \left| \langle \pphiP(\mathbf{X}),   \pphiP(\mathbf{Y}) \rangle   - K(\mathbf{X}, \mathbf{Y}) \right| > \epsilon \right]$ does not exceed the quantity $\frac{\mathcal{C}_\rho}{\nu^3 \epsilon^2} \exp  \left(\tfrac{9  m_4(\mathcal{P}) -  2\sigma^4}{\sigma^8}\right)$.
\end{cor}

This result ensures that the probability of the undesired event $\left| \langle \pphiP(\mathbf{X}),   \pphiP(\mathbf{Y}) \rangle   - K(\mathbf{X}, \mathbf{Y}) \right| > \epsilon$ is small, since upper bounded by a quantity which is inversely quadratic in $\epsilon$ and inversely cubic in $\nu$: this means that even a small value of $\nu$ ensures the latter probability to be small and guarantees the soundness of the approximation.


\subsection{An alternative formulation}\label{sez:kron-tr}
If inspecting equation \eqref{eq:fexact}, it would be natural to replace classical exponentiation - which works with scalars - with Kronecker exponentiation $\X^{\otimes n}$. However, with respect to the feature map $\pphiP$ presented in the previous Section, one may observe that $\otimes_{\kappa = 1}^n {\W^\kk}^\top \X \neq \X^{\otimes n}$ for a general distribution of the weights (the equality would be true only if $\W^\kk$ equals to the identity matrix $\mathbf{I}$ for every $\kappa$). We could thus argue that the following expression would be more appropriate for $\varphi$. 

\begin{defn}\label{def:kronE}
	Using the previous notations, for any $d \times d$ matrix $\X$ we define the scalar quantity
	\begin{equation}\label{eq:kron_exp}
	\kea(\mathbf{X}) = \dfrac{1}{\sigma^{2n}}\sqrt{ 
		\dfrac{\exp(-\tfrac{1}{\sigma^2})}{\rho(n) n!} } {\rm tr}( \mathbf{V}^\top 
	\X^{\otimes n})
	\end{equation}
	where $n \sim \rho$, we still require $\kea$ to satisfy Assumption A.2 (see Theorem \ref{thm:kronP-Exp}), while also assuming
	\begin{enumerate}[{A$'$}.1]
		\item The matrix $\mathbf{V}$ is the Kronecker product of $n$ matrices of size $d \times d$, whose entries are drawn independently from 	$\mathcal{N}(0,\sigma^2)$ (so are consequently the entries of $\mathbf{V}$ -- see suppl. material).
	\end{enumerate}\begin{enumerate}[{A}.2]
	\item The $d \times d$ matrix which is inputted to $\kpa$ lies on the Frobenius norm-unitary sphere, that 
	is $\| \X \|_F = 1.$ 	
	\end{enumerate}
	Then, define the $\nu$ dimensional vector $\pphiE(\X)$ where each component is an independent realization of $\kea(\X)/\sqrt{\nu}$. The explicit steps to compute $\pphi_{\sf{kron}-{\displaystyle e}}(\X)$ given $\X$ are enumerated in Algorithm \ref{alg:kron_exp}.
\end{defn}

At a first glance, equation \eqref{eq:kron_exp} seems closer to an arbitrary component of the exact feature map \eqref{eq:fexact}. This is because, as opposed to \eqref{eq:varphi_kron-P}, the exponentiation operator for scalars is here directly replaced with the Kronecker exponentiation for matrices. Again, as for $\pphi_{\sf{kron}-{\displaystyle \pi}}$, we introduce some random weights -- here, denoted by $\mathbf{V}$ in order to accommodate for the compression generated by approximating an infinite dimensional vector. 

For what concerns the assumptions, A.2 was also hypothesized in Section \ref{sez:proofs} and can be considered as a simple pre-processing step where each entry of the data $\mathbf{X}$ is divided by $\| \X \|_F.$ On the contrary, if we compare A.1 with A'.1, we find a remarkable difference. In fact, A.1 was only constraining the mean and variance of the distribution $\mathcal{P}$. 
Differently, A'.1 not only constrains the probability distribution to be Gaussian but, additionally, we have to explicitly assume that $\mathbf{V}$ factorizes as the Kronecker product of $n$ variables. Indeed, despite $\pphiE$ seems more naturally close to the exact feature map than $\pphiP$, it needs the more restrictive assumption A$'$.2. Without the latter, it is impossible to prove any theoretical result about the approximation $\langle \pphiE(\X),\pphiE(\Y) \rangle$ for \eqref{eq:K}. The reason for that is extremely technical and we illustrate it in the Supplementary Material, Section 4.

\begin{algorithm}[t!]
	\KwIn{A $d \times d$ input matrix $\mathbf{X}$, the desired feature size 
		$\nu$, the 
		probability distributions $\rho$ over integers and $\mathcal{P}$ over real 
		numbers, the kernel bandwith $\sigma > 0$.}
	\KwOut{$		[\phi_{\sf{kron}-{\displaystyle e},1}(\X),\dots,\phi_{\sf{kron}-{\displaystyle e},\nu}(\X)]$} 
	\ForEach{$j = 1,\dots,\nu$}{
		\nl Sample $n$ according to $\rho$\\
		\nl Sample $\mathbf{V}$ as the Kronecker product of $n$ random $d \times d$ matrices, each of the independently sampled from $\mathcal{P}$ \;
		\nl Compute the scalar $e(\X) = \tr\left( 
		\mathbf{V}^\top \X^{\otimes n} \right)$ \\
		\nl {\bf Return} $\phi_{\sf{kron}-{\displaystyle e},j}(\X) = \sigma^{-2n} 
		\left( \tfrac{\exp(-\sigma^{-2})}{\nu \rho(n) n!} \right)^{1/2} 
		e(\X)$}
	\caption{Approx, by Kronecker product}
	\label{alg:kron_exp}
		\vspace{.3 cm}
\end{algorithm}

It is straightforward to see that Definition \ref{def:kronE} actually corresponds to the generalization to the kernel \eqref{eq:K} of the approach in \cite{ker_approx}, which is instead explicitly devised for the log-Euclidean kernel of covariance operators. Here, in fact, $\X$ and $\Y$ can be generic $d \times d$ data structures. Ultimately, we can state that the approximation devised in \cite{ker_approx} is a particular case of $\pphiE$, which, in turn, is a reformulation of $\pphiP$. We can also prove what follows.

\begin{thm}[Unbiased approximation and bound on variance for 
	$\pphi_{\sf{kron}-{\displaystyle e} }$]\label{thm:kronE-Exp}
	Under the assumptions A$'$.1 and A.2, the linear kernel $\langle 
	\pphi_{\sf{kron}-{\displaystyle e} }(\X),\pphi_{\sf{kron}-{\displaystyle e} }(\Y)\rangle$ 
	induced by $\kpa$ is an unbiased estimator for $K(\X,\Y) = 
	\exp(-\tfrac{1}{\sigma^2}\| \X - \Y \|_F^2)$. Actually it results
	\begin{equation}
	\mathbb{E}_{n,\mathbf{V}} \left[ \langle 
	\pphi_{\sf{kron}-{\displaystyle e} }(\X),\pphi_{\sf{kron}-{\displaystyle e} }(\Y)\rangle \right] = 
	K(\X,\Y),
	\end{equation}
	being the expected value jointly computed over all possible realizations of $n$ 
	from 
	$\rho$ and of the weight matrix $\mathbf{V}$. 
	
	In addition, the variance of the proposed estimator is explicitly bounded according to the following inversely-cubic function of $\nu$, for $C_\rho$ as in \eqref{eq:Crho},
	\begin{equation}\nonumber
	\mathbb{var}_{n,\mathbf{V}} \left[ \langle \pphi_{\sf{kron}-{\displaystyle e}}(\X), \pphi_{\sf{kron}-{\displaystyle e}}(\Y) \rangle \right] \leq \dfrac{C_\rho}{\nu^3} \exp\left( \dfrac{3-2\sigma^2}{\sigma^4} \right). 
	\end{equation}
	
	As a corollary, for any $\X,\Y$ and $\epsilon > 0$, the probability $	\mathbb{P}\left[ |\langle \pphiE(\X),\pphiE(\Y) - K(\X,\Y) \rangle| > \epsilon \right]$ is less or equal than $\tfrac{C_\rho}{\nu^3 \epsilon^2} \exp\left( \tfrac{3-2\sigma^2}{\sigma^4} \right)$.
\end{thm}

\proof See the Supplementary Material, Section 4. \endproof

\subsection{The perceptron heuristics}\label{sez:MLPh}
So far, $n$ was randomly sampled from the distribution $\rho$. However, for both $\pphiE$ and $\pphiP$, sampling big values of $n$ increases the number of Kronecker products to be computed and this impact on the computational cost of the method which will be discussed in Section \ref{sez:cost}.

In this Section, we want to investigate the case where, in order to circumvent the previous issue, we fix $n = 1$ in a deterministic manner. This makes \eqref{eq:varphi_kron-P}  and \eqref{eq:kron_exp} formally identical and corresponds selecting only the component of degree 1 in \eqref{eq:fexact}. In these terms, we can interpret it as a \textit{linearization} of the exact feature map associated to the RBF kernel function \eqref{eq:K}. 

Intuitively, the randomness in $n$ can lead to ``explore'' all the infinite components in the exact feature map $\f(\X)$ in order to accumulate enough patterns in $\pphiE$ and $\pphiP$ to properly  approximate the RBF Gaussian kernel. Imposing $n=1$ can be instead thought of as a sort of linearization as to approximate $\f(x)$ in \eqref{eq:fexact}. In such a case, there is clearly a little room for the weights $\W^\kk$ to help recovering from the compression. Therefore, as an opposed paradigm to randomly sample the weights, we can try to learn them in a data-driven fashion, in order to promote class-disambiguation. In fact, since our ultimate goal is accomplishing the action recognition task, the perspective of learning from the data itself seems appealing, especially due to the recent outstanding performance of (deep) feature learning methods \cite{Shahroudy:CVPR16,Liu:ECCV16,JCNN1,JCNN2,Ke:CVPR17,Liu:CVPR17,kermeetfeat}. 

Motivated by the previous considerations, we are now interested in \textit{learning} the weights of $\kpa$ from data. We propose to do so by taking advantage of the formal analogy between $\kpa$ and the hidden layer of a perceptron. Since $n = 1$, we only have $\mathbf{W}^{(1)} = \mathbf{W}$ in \eqref{eq:varphi_kron-P} and we can also write
\begin{equation}
\kpa(\X) \propto tr({\mathbf{W}}^\top \X) = \langle \mathbf{W}, \X \rangle_F = {\rm vec}(\mathbf{W})^\top {\rm vec}(\X).
\end{equation}

As a result, if we denote as $\pmb{\mathbb{W}}$ the $\nu \times d^2$ matrix which stacks by rows all the parameters $\mathbf{W} = \mathbf{W}^{(1)}$ of each independent realization of $\kpa$, we get that 
\begin{equation}\label{eq:hidden}
\pphi_{\sf{kron}-{\displaystyle \pi}}(\X) = \pmb{\mathbb{W}} {\rm vec}(\X)^\top
\end{equation}
meaning that $\pphi_{\sf{kron}-{\displaystyle \pi}}$ actually computes the hidden representation of a (1-layer) perceptron fed with (the vectorization of) $\X$ as data. Furthermore, a squeezing non-linearity (such as $\tanh$ or sigmoid) function on top of \eqref{eq:hidden} can be actually interpreted as a sort of data normalization which is a good practice before SVM training. Since the latter can be implemented in a neural network by means of a hinge loss with weight decay, we can therefore establish a connection between our paradigm $\pphi_{\sf{kron}-{\displaystyle \pi}}$ + linear SVM and a feed-forward perceptron, having one hidden layer of size $\nu$, with sigmoid non-linearities and hinge loss with weight decay for final classification. 

Let us summarize the previous findings.
Consider $\pphi_{\sf{kron}-{\displaystyle \pi}}$, set $n = 1$ and, instead of a random sampling, learn the weights $\mathbf{W} = \mathbf{W}^{(1)}$ for each $\kpa$-component from the hidden layer of the architecture composed by a supervised feed-forward perceptron with sigmoid as non-linearities and cross entropy loss. Then, use the network to extract the feature map, that we term $\pphi_{\rm P}$, and use it in combination of a linear SVM. This can be interpreted as a deterministic implementation of $\kea$ and $\kpa$ where random weights' sampling is replaced with their data-driven optimization. (see the Supplementary Material, Section 5, for further details).

In Section \ref{sez:soa}, we will validate the previous heuristics of replacing $\pphiP$ as given by Algorithm \ref{alg:kronP} with the map $\pphi_{\rm P}$ which is computed according to pseudo-code presented in Algorithm \ref{alg:MLP}.

\begin{algorithm}[t!]
	\KwIn{A $d \times d$ input matrix $\mathbf{X}$, a training set $\boldsymbol{\mathcal{D}}$ of $d \times d$ matrices, the desired feature size 
		$\nu$, the 
		probability distributions $\rho$ over integers and $\mathcal{P}$ over real 
		numbers, the kernel bandwith $\sigma > 0$.}
	\KwOut{The $\nu$-dim feature map $\pphi_{\rm P}(\X)$} 
		\nl Learn $\nu \times d^2$ weight matrix $\pmb{\mathbb{W}}$ from the hidden layer parameters of the architecture of \cite{kermeetfeat} trained on $\boldsymbol{\mathcal{D}}$. \\
		\nl {\bf Return} $\pphi_{\rm P}(\X)$ as the multiplication of $\pmb{\mathbb{W}}$ by the vectorization of $\X$. \\
	\caption{The perceptron heuristics.}
	\label{alg:MLP}
		\vspace{.3 cm}
\end{algorithm}

\section{Evaluation vs. other approximations}\label{sez:exp}

In this Section we will present our experimental validation of $\pphi_{\sf{kron}-{\displaystyle \pi}}$, $\pphi_{\sf{kron}-{\displaystyle e}}$ as well as the perceptron heuristics $\pphi_{\rm P}$. To begin with, we will describe the benchmark datasets adopted and explain the data preprocessing that we carried out.

\subsection{3D action recognition datasets and preprocessing}

We present here all the datasets considered for the experiments, namely UTKinect \cite{UTKinect}, Florence3D \cite{Florence3D}, MSR-Action-Pairs (MSR-$pairs$) \cite{MSRPairs}, MSR-Action3D \cite{Action3D}, Gaming-3D (G3D) \cite{G3D}, HDM-05 \cite{HDM-05}, MSRC-Kinect12 \cite{MSRC} and NTU RGB+D \cite{Shahroudy:CVPR16}. Table \ref{tab:tab:tab} summarizes statistics for each. 

We follow usual training and testing splits proposed in the literature. For Florence3D, G3D, and UTKinect, we use the protocols of \cite{Vemulapalli:CVPR14,Vemulapalli:CVPR16,Camps:CVPR16}. For MSR-Action3D, we adopt the splits originally proposed by \cite{Action3D}. On MSRC-Kinect12, once highly corrupted action instances are removed as in \cite{egizi}, training is performed on odd-index subject, while testing on the even-index ones. On HDM-05, the training split exploits all the data from the ``\texttt{bd}'' and ``\texttt{mm}'' subjects, being  ``\texttt{bk}'', ``\texttt{dg}'' and ``\texttt{tr}'' left out for testing \cite{Wang:ICCV15}. To be consistent with the literature, we replicated the 14 classes experiments (HDM-05$_{14}$) as in \cite{Wang:ICCV15,Cavazza:ICPR16}. When dealing with the whole dataset (HDM-05$_{all}$), since some of the total classes are missing from the training/testing splits, we adopted the protocol of \cite{bellolui} to partition the dataset into 65 action classes. For NTU RGB+D, we followed the authors' instruction \cite{Shahroudy:CVPR16} in removing the most corrupted instances, also purging the trials with missing joints recordings. Finally, we replicated both the cross-subject and cross-view testing protocols proposed in \cite{Shahroudy:CVPR16}, denoting them as NTU-$\times$-$subject$ and NTU-$\times$-$view$.

In all experiments, as a common data pre-processing step \cite{Vemulapalli:CVPR14,Camps:ACCV14,Vemulapalli:CVPR16,Camps:CVPR16,Shahroudy:CVPR16,Cavazza:ICPR16,ECCV16,Liu:ECCV16}, we fix one root joint (the one located at the hip center), and we compute the relative differences of all the other $J-1$ 3D joint positions. By doing this at any timestamps $t = 1,\dots,T$ we obtain a $3(J-1)$-dimensional (column) vector $\mathbf{p}(t)$ of relative displacements. As the representation for data instance $[\mathbf{p}(1),\dots,\mathbf{p}(T)]$, we compute a covariance matrix
\vspace{-.3 cm}
\begin{equation}\label{eq:cov}
\mathbf{C} = \dfrac{1}{T - 1} \sum_{t = 1}^T (\mathbf{p}(t) - \boldsymbol{\mu})(\mathbf{p}(t) - \boldsymbol{\mu})^\top,
\vspace{-.1 cm}
\end{equation}
being $\boldsymbol{\mu} = \frac{1}{T} \sum_{t = 1}^T \mathbf{p}(t)$ the temporal average of $\mathbf{p}(t)$. Finally, the input representation for our approximated feature map is obtained as
\begin{equation}
\X = \log \mathbf{C} = \mathbf{U} {\rm diag}(\log(\boldsymbol{\varsigma} )) {\mathbf{U} }^\top,
\end{equation}
being $\boldsymbol{\varsigma}$ the vector of eigenvalues (eventually regularized by an additive factor as in \cite{Minh:NIPS14}) and $\mathbf{U} $ the matrix of eigenvectors of $\mathbf{C}$. Finally, since the $\log$ of a symmetric matrix is symmetric, in order to avoid to process identical entries twice, we zero out all the lower triangular entries in $\X$ and we divide all of them by $\| \X \|_F$.

\begin{table}[t!]
	\footnotesize
	\centering
	\begin{tabular}{|r|c|c|c|c|c|}
		\hline
		& {\tiny Classes} & {\tiny Subjects} & {\tiny Repetitions} & {\tiny Samples} & {\tiny Joints} \\\hline\hline
		UTKinect \cite{UTKinect} & 10 & 10 & 1-2 & 199 & 20 \\
		Florence3D \cite{Florence3D} & 9 & 10 & 2-3 & 215 & 15 \\
		MSR-$pairs$ \cite{MSRPairs} & 20 & 10 & 1-3 & 353 & 20 \\
		MSR-Action3D \cite{Action3D} & 20 & 10 & 2-3 & 567 & 20 \\
		G3D \cite{G3D} & 20 & 10 & 2-4 & 663 & 20 \\		
		HDM-05$_{14}$ \cite{HDM-05} & 14 & 5 & 8-10 & 686 & 31 \\ 
		HDM-05$_{all}$ \cite{HDM-05} & 65 & 5 & 8-40 & 2343 & 31 \\
		MSRC-Kinect12 \cite{MSRC} & 12 & 30 & 10-25 & 5881 & 20 \\
		NTU-$\times$$subject$ \cite{Shahroudy:CVPR16} & \multirow{2}{*}{60} & \multirow{2}{*}{40} & \multirow{2}{*}{80} & \multirow{2}{*}{56578} & \multirow{2}{*}{25} \\
		NTU-$\times$$view$ \cite{Shahroudy:CVPR16} & & & & & \\\hline
	\end{tabular}
	\vspace{.3 cm}
	\caption{\normalsize Statistics about the 3D action recognition datasets 
	considered.}
	\label{tab:tab:tab}
\end{table}

\subsection{Implementation details}

The implementation for $\pphiP$, $\pphiE$ and $\pphi_{\rm P}$, is in MATLAB, and is based on the pseudo-code of Algorithms \ref{alg:kronP}, \ref{alg:kron_exp} and \ref{alg:MLP}, respectively\footnote{We used MATLAB R2017a installed on an Intel Xeon(R) CPU E5645 @2.40GHz, 12 cores, with 12GB RAM.}. 

For both $\pphiP$ and $\pphiE$, we fixed the distribution $\rho$ to be a Geometric with parameter $0.9$ - a full justification fro this choice is provided in Section \ref{sez:cost} - and $\mathcal{P} = \mathcal{N}(0,\sigma^2)$ is a Gaussian distribution. We carried out experiments for $\nu=\textrm{10, 20, 50, 100, 200, 500, 1000, 2000, 5000}$ and we performed 10 repetitions of each experiment, to account for the random nature of the approach. Action recognition performance is thus the \textit{mean} classification accuracy over the 10 runs.

For $\pphi_{\rm P}$ we used the architecture of \cite{kermeetfeat} which is also fed with $\log$-projected covariance representations. With such input we trained a one-hidden layer perceptron with sigmoid non-linearities and cross-entropy loss using scaled conjugate gradient descent for all datasets except to the NTU RGB+D, for which we used ADAM optimizer with mini-batches of size 1024. The size of the hidden layer was cross-validated among $10^2,10^3,\dots,10^9$ as the one which gives the lowest objective value. Once the network is trained, we extracted the parameter of the hidden layer and built $\pphi_{\rm P}$ as in Algorithm \ref{alg:MLP}. For all cases - $\pphiP$, $\pphiE$ and $\pphi_{\rm P}$ - we use the linear SVM implementation of \cite{liblinear}. 

\subsection{Compared methods}

We compare our approach with the following approximating schemes proposed in the literature.


We consider the Fourier based approach, originally proposed in \cite{RR:NIPS07} and afterwards extended in \cite{Vedaldi:BMVC,Vedaldi_tPAMI}. We also considered the two alternative approximations of \cite{Fastfood} and \cite{KK:AISTATS13}, based on Hadamard transform and Taylor expansion, respectively.
We did not compare with \cite{Ring:PRL16} because, despite the work is related to our case, the cost of the proposed approximation is prohibitive, being exponential with respect to the input data dimensionality.

{\bf State-of-the-art approaches for 3D action recognition.} As mentioned in Section \ref{sez:RW}, action recognition from skeletal joints has experienced a tremendous interest from the community and, currently, the state-of-the-art performance is disputed against two classes of powerful approaches: kernel methods based on covariance or rotation matrices and feature learning techniques which exploit deep (and often recurrent) neural networks. Since we consider many competitors, which are different from dataset to dataset, we will comment on each of them at the moment we present the results obtained by $\pphi_{\rm P}$ (see Section \ref{sez:soa}).

\subsection{Comparing the bounds on the variance}

A direct comparison among the bounds of the variance between the proposed approximations $\pphiP,\pphiE$ and the previous approaches \cite{RR:NIPS07,Vedaldi:BMVC,Vedaldi_tPAMI,Fastfood,KK:AISTATS13,Ring:PRL16,ker_approx} is tricky because the theoretical foundation of each approximation is approached in different manners. Indeed, \cite{Ring:PRL16} does not rely on a probabilistic framework, but instead, proposes a simple truncation of the feature map \eqref{eq:fexact}. Despite this allows the relative error between the exact and the approximated kernel to be explicitly bounded, the method \cite{Ring:PRL16} is only applicable in a case of a small $d$ (see Section \ref{sez:cost}). 

Other probabilistic frameworks as \cite{RR:NIPS07,Vedaldi:BMVC,Vedaldi_tPAMI,KK:AISTATS13} also provide an analogous result of Corollary \ref{cor:kronP}. However, the probability $\mathbb{P}[| \langle \pphi(\X),\pphi(\Y)\rangle - K(\X,\Y) | > \epsilon ]$ has only a weaker $O(1/(\nu\epsilon^2))$ behavior. Moreover, while considering the analogous approximated feature maps of \cite{RR:NIPS07,Vedaldi:BMVC,Vedaldi_tPAMI,KK:AISTATS13}, the previous result holds only provided that  $\X$ and $\Y$ lie on a common sub-manifold. Despite we analogously assume that $\X$ and $\Y$ have unitary norm, our requirement is easier to satisfy in practice and less restrictive. Moreover, ancillary conditions are needed in those works to achieve results of the form $\mathbb{P}[| \langle \pphiP(\X),\pphiP(\Y) - K(\X,\Y) \rangle| > \epsilon ]$ and $\mathbb{P}[| \langle \pphiE(\X),\pphiE(\Y) - K(\X,\Y) \rangle| > \epsilon ]$, while, differently, the assumptions we made are much milder.

In addition, \cite{Fastfood} and \cite{ker_approx} also provide a strong theoretical foundation similar to ours. Indeed, in \cite{Fastfood}, the proposed approximation gives an unbiased estimation of \eqref{eq:K} and its variance is bounded a $O(1/\nu)$ function. In our case, however, the variances of the approximations $\pphiP$ and $\pphiE$ are bounded by $O(1/\nu^3)$ and are therefore more rapidly decreasing to zero, ensuring a better approximation for a fixed $\nu$. Finally, the quality of the approximation of \cite{ker_approx} is comparable - no bias and $O(1/\nu^3)$ decreasing variance. This is reasonable because, as we proved, 
\cite{ker_approx} is a particular case of our approximation $\pphiP$.

\subsection{Computational cost}\label{sez:cost}

Interestingly, we can observe one common trend which is shared across all the approaches \cite{RR:NIPS07,Vedaldi:BMVC,Vedaldi_tPAMI,KK:AISTATS13,ker_approx}: in computational terms, the number of products required for computing one component of the feature map is linear with respect to the data dimensionality (which is $O(d^2)$ since $\log$-covariance $d \times d$ matrices are used as input). Among the previously published works, two papers are different: \cite{Fastfood} achieves a log-linear complexity, while, unfortunately, \cite{Ring:PRL16} has exponential complexity with respect to the data size: this is the reason why we were not able to include \cite{Ring:PRL16} among the methods in comparison.

We can thus observe that the cost of calculating $\tr(\otimes_{\kappa = 1}^n {\mathbf{W}^\kk}^\top \mathbf{X}) = \prod_{\kappa = 1}^{n}\tr( {\mathbf{W}^\kk}^\top \mathbf{X})$ (in the computation of $\pphiP$) is linear in both the input data dimensionality and in $n$. Similarly, the same holds for  $\pphiE$, thanks to the factorization assumption A$'$.1.

Despite such linear dependence from $n$ may appear as a drawback, we can take advantage of the freedom in choosing $\rho$ in order to keep $n$ small. Indeed, throughout all the experiments, either involving $\pphiP$ or $\pphiE$, we fixed $\rho$ as a Geometrical distribution of parameter $\theta = 0.9$. This ensures that the probability of sampling high values of $n$ from $\rho$ is practically zero. Indeed, through analytical computations, we can also notice that, for each realization of $\kpa$ or $\kea$, $\mathbb{P}(n>3)=0.04$. 

This makes the computational cost of our approach substantially in line with that of other works\cite{RR:NIPS07,Vedaldi:BMVC,Vedaldi_tPAMI,KK:AISTATS13,ker_approx}.



\subsection{Analysis of action recognition performance}\label{sez:against}

\begin{figure*}
			\vspace{.3 cm}
	\centering
	\tikzset{every mark/.append style={scale=1.5}}
	\tikzset{every axis title/.style={below right,at={(.56,.63)}}}
	\hspace{-9 pt}
	\begin{tikzpicture}
	\begin{axis}[
	width=.49\columnwidth,
	title={\bf HDM-05$_{\boldsymbol{all}}$ \cite{HDM-05}},
	xlabel={Feature dimension [$\nu$]},
	ylabel={Classification accuracy [\%]},
	xmin=0.5, xmax=9.5,
	ymin =  8, ymax = 70,
	ytick={10,20,30,50,70},
	xtick={1,2,3,4,5,6,7,8,9},
	xticklabels = {10, 20, 50, 100, 200, 500, 1000, 2000, 5000},
	xticklabel style={rotate=-40,anchor=north west},
	legend pos=north west,
	ymajorgrids=true,
	grid style=dashed,
	]
	\addplot[color=brown,mark=*,line width = 5 pt]
	coordinates{(9,64.01)};
	
	\addplot[color=black, loosely dashed, line width = 2 pt]
	coordinates{(.5,65.90)(9.5,65.90)};
	\addplot[color=green,mark=triangle*, line width = 1.5 pt]
	coordinates{(1,12.09)(2,17.59)(3,26.98)(4,36.78)(5,47.62)(6,60.38)(7,63.21)(8,63.66)(9,62.98)};
	
	\addplot[color=magenta,mark=pentagon*, line width = 2 pt]
	coordinates{(1,8.96)(2,13.28)(3,21.55)(4,28.71)(5,37.17)(6,43.9)(7,50.4)(8,58.69)(9,62.01)};
	\addplot[color=blue,mark=diamond*,line width = 1.3 pt]
	coordinates{(1,21.16)(2,29.42)(3,32.32)(4,46.01)(5,54.99)(6,60.98)(7,61.42)(8,62.84)(9,63.98)};
	\addplot[color=orange,mark=square*]
	coordinates{(1,10.72)(2,18.02)(3,31.91)(4,47.5)(5,57.66)(6,63.42)(7,65.4)(8,66.51)(9,67.5)};
	\end{axis}
	\end{tikzpicture}\hspace{8 pt}
	\begin{tikzpicture}
	\begin{axis}[
	width=.49\columnwidth,
	title={\bf MSRC-Kinect12 \cite{MSRC}},
	xlabel={Feature dimension [$\nu$]},
	xmin=0.5, xmax=9.5,
	ymin = 49, ymax = 96,
	ytick={49,60,70,80,90},
	xtick={1,2,3,4,5,6,7,8,9},
	xticklabels = {10, 20, 50, 100, 200, 500, 1000, 2000, 5000},
	xticklabel style={rotate=-40,anchor=north west},
	ymajorgrids=true,
	grid style=dashed,
	legend pos=south east,
	]
	
	\addplot[color=black, loosely dashed, line width = 2 pt]
	coordinates{(.5,94.30)(9.5,94.30)};\addlegendentry{Exact kernel \eqref{eq:K}}
	
	\addplot[color=brown,mark=*,line width = 4 pt]
	coordinates{(8,92.23)(9,92.21)};\addlegendentry{Hadamard \cite{Fastfood}}
	
	\addplot[color=green,mark=triangle*, line width = 2.3 pt]
	coordinates{(1,68.71)(2,80.07)(3,86.89)(4,87.45)(5,89.2)(6,91.32)(7,92.23)(8,92.26)(9,92.4)};\addlegendentry{Fourier \cite{RR:NIPS07,Vedaldi:BMVC,Vedaldi_tPAMI}}
	
	\addplot[color=magenta,mark=pentagon*, line width = 2 pt]
	coordinates{(1,49.87)(2,60.71)(3,71.72)(4,78.03)(5,81.28)(6,83.7)(7,92.59)(8,92.81)(9,92.03)};\addlegendentry{Taylor \cite{KK:AISTATS13}}
	
	\addplot[color=blue,mark=diamond*,line width = 1.3]
	coordinates{(1,70.36)(2,82.32)(3,89.33)(4,90.95)(5,91.92)(6,92.28)(7,92.23)(8,92.34)(9,92.36)};\addlegendentry{$\pphi_{\sf{kron}-{\displaystyle e}}$ \emph{(proposed)} \cite{ker_approx}}
	
	\addplot[color=orange,mark=square*]
	coordinates{(1,69.58)(2,83.11)(3,89.45)(4,91.06)(5,91.92)(6,92.75)(7,93.45)(8,94.59)(9,95.63)};\addlegendentry{$\pphi_{\sf{kron}-{\displaystyle \pi}}$ \emph{(proposed)}}
	
	\legend{{\footnotesize Exact RBF kernel machine}, {\footnotesize Hadamard approx \cite{Fastfood}},{\footnotesize Fourier approx \cite{RR:NIPS07,Vedaldi:BMVC,Vedaldi_tPAMI}},{\footnotesize Taylor approx \cite{KK:AISTATS13}},{\footnotesize $\pphi_{\sf{kron}-{\displaystyle e}}$ \emph{(proposed)}},{\footnotesize $\pphi_{\sf{kron}-{\displaystyle \pi}}$ \emph{(proposed)}}}
	
	\end{axis}
	\end{tikzpicture}
	\caption{\normalsize Medium data regime ($\sim10^3$ samples). Experiments 
	on the full HDM-05 datasets and on the MSRC-Kinect 12. In each case, we 
	monitor the changes in action recognition accuracy as a function of the feature 
	dimensionality $\nu$. Across figures, the same color refers to same method. 
	Best viewed in color.}\label{fig:medium}
\end{figure*}
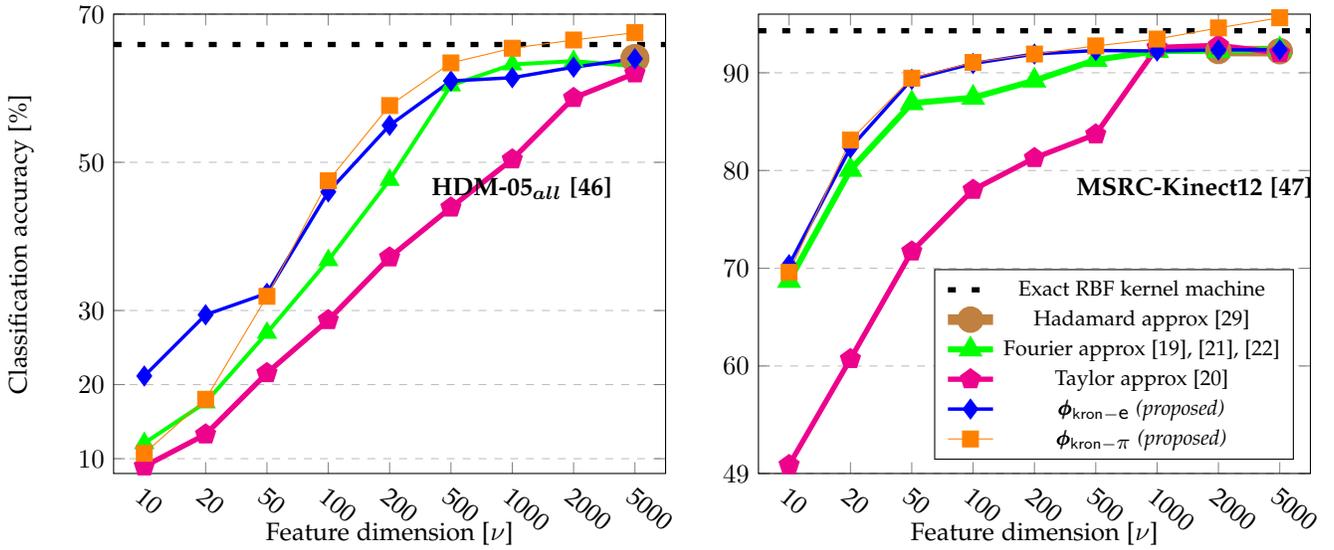

Despite the several approximations \cite{RR:NIPS07,Vedaldi:BMVC,Vedaldi_tPAMI,Fastfood,KK:AISTATS13,Ring:PRL16,ker_approx} have been proposed and are applicable to a RBF kernel function \eqref{eq:K}, to the best of our knowledge there is no clear evidence of which method is effective and efficient for classification. Indeed, despite all methods ensure scalability in the big data regime, there is no clear understanding about which method gives superior performance and, in general, how a good feature dimensionality $\nu$ should be chosen in practice. Here, we try to answer this question with a detailed analysis of 3D action recognition accuracies on the 10 benchmark datasets listed in Table \ref{tab:tab:tab}, while the feature dimensionality $\nu$ assumes one of the following values: 10, 20, 50, 100, 200, 500, 1000, 2000, 5000. We report the results of this analysis in Figures \ref{fig:small}, \ref{fig:medium} and \ref{fig:big} and we will discuss the scored results in the following.

\begin{figure*}
			\vspace{.3 cm}
	\centering
	\tikzset{every mark/.append style={scale=1.5}}
	\tikzset{every axis title/.style={below right,at={(.56,.63)}}}
	\tikzset{every legend style/.style={at={(0.45,0.5)}}}
	\hspace{-9 pt}
	\begin{tikzpicture}
	\begin{axis}[
	name = UT,
	width=.49\columnwidth,
	title={\bf UTKinect \cite{UTKinect}},
	ylabel={Classification accuracy [\%]},
	xmin=0.5, xmax=9.5,
	ymin=48.6, ymax=85.2,
	ytick={40,50,60,70,80},
	xtick={1,2,3,4,5,6,7,8,9},
	xticklabels = {},
	legend pos=north west,
	ymajorgrids=true,
	grid style=dashed,
	]
	\addplot[color=brown,mark=*,line width = 5 pt]
	coordinates{(8,84)(9,83.61)};
	
	\addplot[color=black, loosely dashed, line width = 2 pt]
	coordinates{(.5,85)(9.5,85)};
	\addplot[color=green,mark=triangle*, line width = 1.5 pt]
	coordinates{(1,50.6)(2,53.36)(3,60.25)(4,65.93)(5,74.07)(6,80.89)(7,81.63)(8,83.5)(9,83.9)};
	
	\addplot[color=magenta,mark=pentagon*, line width = 2 pt]
	coordinates{(1,48.88)(2,59.52)(3,70.57)(4,76.87)(5,80.28)(6,82.92)(7,83.91)(8,84.05)(9,83.21)};
	\addplot[color=blue,mark=diamond*,line width = 1.3 pt]
	coordinates{(1,53.72)(2,67.92)(3,73.77)(4,78.9)(5,81.11)(6,82.31)(7,82.76)(8,82.98)(9,84.44)};
	\addplot[color=orange,mark=square*]
	coordinates{(1,54.97)(2,65.62)(3,75.34)(4,79.47)(5,81.93)(6,82.78)(7,83.43)(8,83.81)(9,84.07)};
	\end{axis}
	\end{tikzpicture}\hspace{8 pt}
	\begin{tikzpicture}
	\begin{axis}[
	name = Flo,
	at=(UT.right of south east), anchor=left of south west,
	width=.49\columnwidth,
	title={\bf Florence3D \cite{Florence3D}},
	xmin=0.5, xmax=9.5,
	ymin=40, ymax=87,
	ytick={40,50,60,70,80},
	xtick={1,2,3,4,5,6,7,8,9},
	xticklabels = {},
	legend pos=north west,
	ymajorgrids=true,
	grid style=dashed,
	]
	
	\addplot[color=brown,mark=*,line width = 2 pt]
	coordinates{(7,81.52)(8,83.16)(9,83.43)};
	
	\addplot[color=black, loosely dashed, line width = 2 pt]
	coordinates{(.5,86.2)(9.5,86.2)};
	
	\addplot[color=green,mark=triangle*,line width = 2 pt]
	coordinates{(1,52.48)(2,67.16)(3,71.5)(4,76.12)(5,81.23)(6,84.74)(7,85.45)(8,85.5)(9,83.8)};

	\addplot[color=magenta,mark=pentagon*, line width = 2 pt]
	coordinates{(1,43.09)(2,55.36)(3,67.64)(4,73.37)(5,78.2)(6,82.15)(7,83.85)(8,83.8)(9,84.15)};
	\addplot[color=blue,mark=diamond*,line width = 1.3 pt]
	coordinates{(1,61.99)(2,72.09)(3,79.36)(4,82.48)(5,82.75)(6,83.68)(7,84.24)(8,84.87)(9,85.9)};
	\addplot[color=orange,mark=square*]
	coordinates{(1,59.6)(2,71.46)(3,78.08)(4,81.23)(5,82.78)(6,83.67)(7,84.3)(8,84.34)(9,85.21)};
	\end{axis}
	\end{tikzpicture}\\
	\hspace{-9 pt}
	\begin{tikzpicture}
	\begin{axis}[
	name = 'A3D',
	at=(UT.below south), 
	width=.49\columnwidth,
	title={\bf MSR-$\boldsymbol{pairs}$ \cite{MSRPairs}},
	ylabel={Classification accuracy [\%]},
	xmin=0.5, xmax=9.5,
	ymin = 36, ymax = 75.2,
	ytick={36,46,56,66,75},
	xtick={1,2,3,4,5,6,7,8,9},
	xticklabels = {},
	legend pos=north west,
	ymajorgrids=true,
	grid style=dashed,
	]
	
	\addplot[color=brown,mark=*,line width = 2 pt]
	coordinates{(8,71.98)(9,72.15)};
	
	\addplot[color=black, loosely dashed, line width = 2 pt]
	coordinates{(.5,74.90)(9.5,74.90)};
	
	\addplot[color=green,mark=triangle*,line width = 2 pt]
	coordinates{(1,37.82)(2,40.74)(3,45.13)(4,50.4)(5,56.66)(6,65.76)(7,71.23)(8,71.1)(9,72.76)};

	\addplot[color=magenta,mark=pentagon*, line width = 2 pt]
	coordinates{(1,36.32)(2,46.96)(3,57.48)(4,63.47)(5,67.64)(6,71.16)(7,72.86)(8,73.58)(9,73.1)};
	\addplot[color=blue,mark=diamond*,line width = 1.3 pt]
	coordinates{(1,37.68)(2,49.42)(3,60.4)(4,65.04)(5,69.08)(6,71.51)(7,71.43)(8,72.38)(9,73.14)};
	\addplot[color=orange,mark=square*]
	coordinates{(1,40.63)(2,50.31)(3,59.82)(4,65.94)(5,69.03)(6,71.91)(7,72.95)(8,73.09)(9,73.6)};
	\end{axis}
	\end{tikzpicture}\hspace{7 pt}
	\begin{tikzpicture}
	\begin{axis}[
	width=.49\columnwidth,
	title={\bf MSR-Action3D \cite{Action3D}},
	xmin=0.5, xmax=9.5,
	ymin = 46, ymax = 90,
	ytick={46,56,66,72,78,84,90},
	xtick={1,2,3,4,5,6,7,8,9},
	xticklabels = {},
	legend pos=north west,
	ymajorgrids=true,
	grid style=dashed,
	]
	
	\addplot[color=brown,mark=*,line width = 2 pt]
	coordinates{(8,87.54)(9,88.15)};
	
	\addplot[color=black, loosely dashed, line width = 2 pt]
	coordinates{(.5,89.40)(9.5,89.40)};
	
	\addplot[color=green,mark=triangle*, line width = 2 pt]
	coordinates{(1,54.48)(2,60.25)(3,67.9)(4,71.28)(5,74.99)(6,81.39)(7,83.68)(8,86.27)(9,88.43)};

	\addplot[color=magenta,mark=pentagon*, line width = 2 pt]
	coordinates{(1,46.1)(2,61.23)(3,76.95)(4,82)(5,86.73)(6,88.56)(7,88.68)(8,89.02)(9,89.64)};
	
	\addplot[color=blue,mark=diamond*,line width = 1.3 pt]
	coordinates{(1,56.82)(2,71.9)(3,81.62)(4,86)(5,87.37)(6,88.67)(7,89.23)(8,89.41)(9,89.47)};
	
	\addplot[color=orange,mark=square*]
	coordinates{(1,56.24)(2,71.65)(3,81.18)(4,86.48)(5,88.69)(6,89.44)(7,89.87)(8,89.81)(9,89.82)};
	\end{axis}
	\end{tikzpicture}\\
	\vspace{8 pt}
	\begin{tikzpicture}
	\begin{axis}[
	width=.49\columnwidth,
	title={\bf G3D \cite{G3D}},
	title style = {at={(.6,.64)}},
	xlabel={Feature dimension [$\nu$]},
	ylabel={Classification accuracy [\%]},
	xmin=0.5, xmax=9.5,
	ymin = 25, ymax = 86,
	ytick={40,50,60,70,80},
	xtick={1,2,3,4,5,6,7,8,9},
	xticklabels={10 ,20 ,50 ,100 ,200 ,500 ,1000 ,2000 ,5000 },
	xticklabel style={rotate=-40,anchor=north west},
	legend pos=south east,
	ymajorgrids=true,
	grid style=dashed,
	]
	
	\addplot[color=black, loosely dashed, line width = 2 pt]
	coordinates{(.5,85.40)(9.5,85.40)};\addlegendentry{Exact kernel \eqref{eq:K}}
	
	\addplot[color=brown,mark=*,line width = 3 pt]
	coordinates{(8,83.01)(9,83.35)};\addlegendentry{Hadamard \cite{Fastfood}}
	
	\addplot[color=green,mark=triangle*, line width = 2.3 pt]
	coordinates{(1,53.71)(2,61.55)(3,67.92)(4,72.34)(5,77.34)(6,81.3)(7,82.85)(8,83.58)(9,83.76)};\addlegendentry{Fourier \cite{RR:NIPS07,Vedaldi:BMVC,Vedaldi_tPAMI}}
	
	\addplot[color=magenta,mark=pentagon*, line width = 2 pt]
	coordinates{(1,25.42)(2,38.48)(3,57.71)(4,70.25)(5,77.72)(6,82.33)(7,83.93)(8,84.55)(9,84.6)};\addlegendentry{Taylor \cite{KK:AISTATS13}}
	
	\addplot[color=blue,mark=diamond*,line width = 1.3]
	coordinates{(1,49.14)(2,62.93)(3,76.46)(4,79.86)(5,81.29)(6,82.94)(7,83.32)(8,83.62)(9,83.85)};\addlegendentry{$\pphi_{\sf{kron}-{\displaystyle e}}$ \emph{(proposed)} \cite{ker_approx}}
	
	\addplot[color=orange,mark=square*]
	coordinates{(1,48.63)(2,63.82)(3,75.74)(4,79.9)(5,82.2)(6,83.43)(7,83.46)(8,83.75)(9,83.87)};\addlegendentry{$\pphi_{\sf{kron}-{\displaystyle \pi}}$ \emph{(proposed)}}
	
	\legend{{\footnotesize Exact RBF kernel machine}, {\footnotesize Hadamard approx \cite{Fastfood}},{\footnotesize Fourier approx \cite{RR:NIPS07,Vedaldi:BMVC,Vedaldi_tPAMI}},{\footnotesize Taylor approx \cite{KK:AISTATS13}},{\footnotesize $\pphi_{\sf{kron}-{\displaystyle e}}$ \emph{(proposed)}},{\footnotesize $\pphi_{\sf{kron}-{\displaystyle \pi}}$ \emph{(proposed)}}}
	
	\end{axis}
	\end{tikzpicture}
	\begin{tikzpicture}
	\begin{axis}[
	width=.49\columnwidth,
	title={\bf HDM-05$_{\boldsymbol{14}}$ \cite{HDM-05}},
	xlabel={Feature dimension [$\nu$]},
	xmin=0.5, xmax=9.5,
	ymin = 29, ymax = 91,
	ytick={40,50,60,70,80},
	xtick={1,2,3,4,5,6,7,8,9},
	xticklabels={10 ,20 ,50 ,100 ,200 ,500 ,1000 ,2000 ,5000 },
	xticklabel style={rotate=-40,anchor=north west},
	legend pos=south east,
	ymajorgrids=true,
	grid style=dashed,
	]
	
	\addplot[color=brown,mark=*,line width = 5 pt]
	coordinates{(9,88.62)};
	
	\addplot[color=black, loosely dashed, line width = 2 pt]
	coordinates{(.5,90.60)(9.5,90.60)};
	
	\addplot[color=green,mark=triangle*, line width = 2 pt]
	coordinates{(1,29.41)(2,33.32)(3,50.21)(4,61.18)(5,73.67)(6,84.02)(7,89.02)(8,89.1)(9,89.21)};
	
	\addplot[color=magenta,mark=pentagon*, line width = 2 pt]
	coordinates{(1,32.98)(2,45.71)(3,66.79)(4,78.16)(5,82.19)(6,84.52)(7,85.11)(8,86.74)(9,88.94)};
	
	\addplot[color=blue,mark=diamond*,line width = 1.3]
	coordinates{(1,44.18)(2,59.18)(3,74.93)(4,80.73)(5,84.73)(6,87.35)(7,88.51)(8,89.07)(9,89.62)};
	
	\addplot[color=orange,mark=square*]
	coordinates{(1,32.12)(2,49.88)(3,70.75)(4,80.82)(5,84.92)(6,88.09)(7,88.92)(8,89.36)(9,89.91)};
	
	\end{axis}
	\end{tikzpicture}
		\vspace{.3 cm}
	\caption{\normalsize Small data regime ($\sim 10^2$ samples). Experiments on 
	the UTKinect, Florence 3D, MSR-$pairs$, MSR-Action3D, G3D and HDM-05 using 
	the selection of 14 classes used by \cite{egizi}. In each case, we monitor the 
	changes in action recognition accuracy as a function of the feature 
	dimensionality $\nu$. Across figures, the same color refers to same method. 
	Best viewed in color.}\label{fig:small}
\end{figure*}
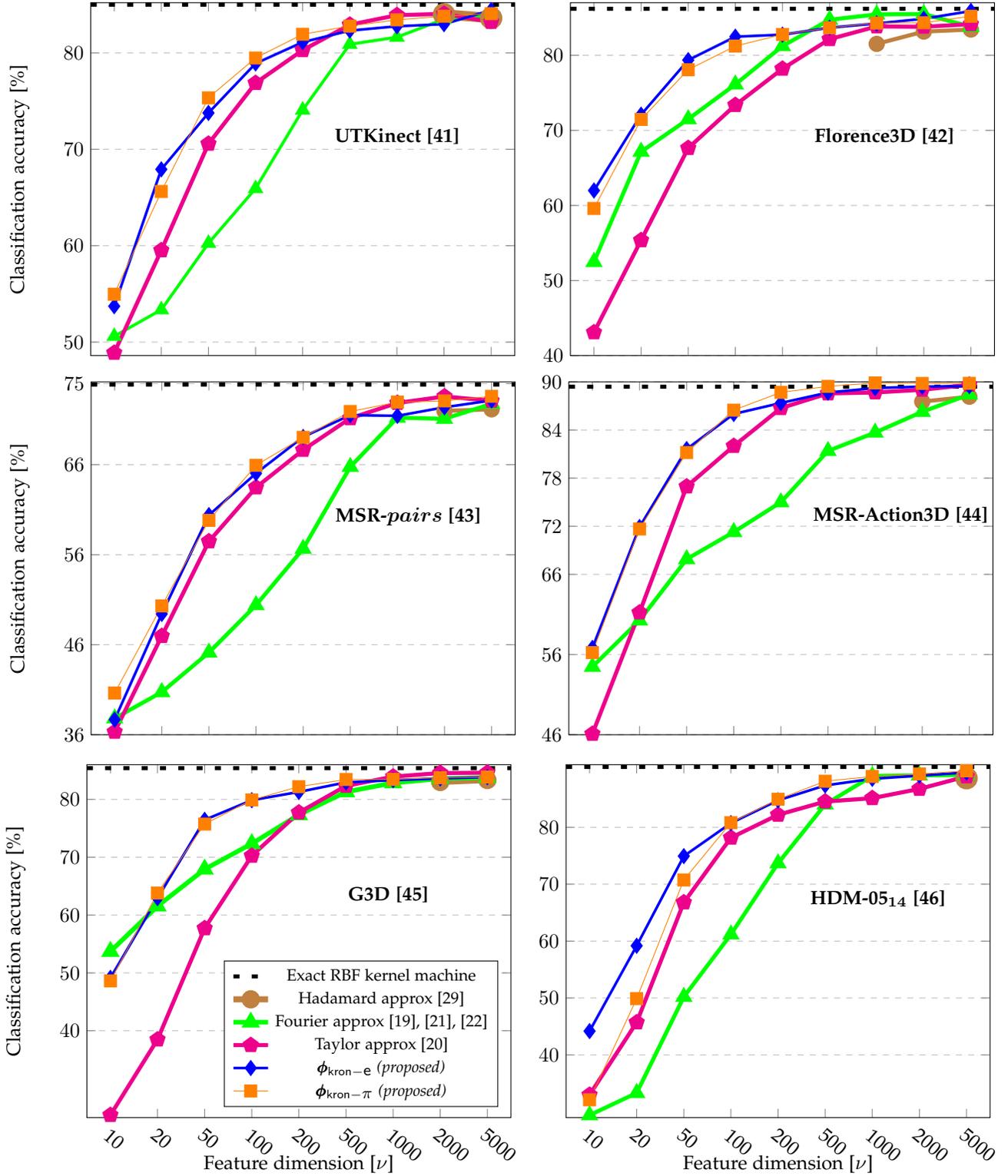

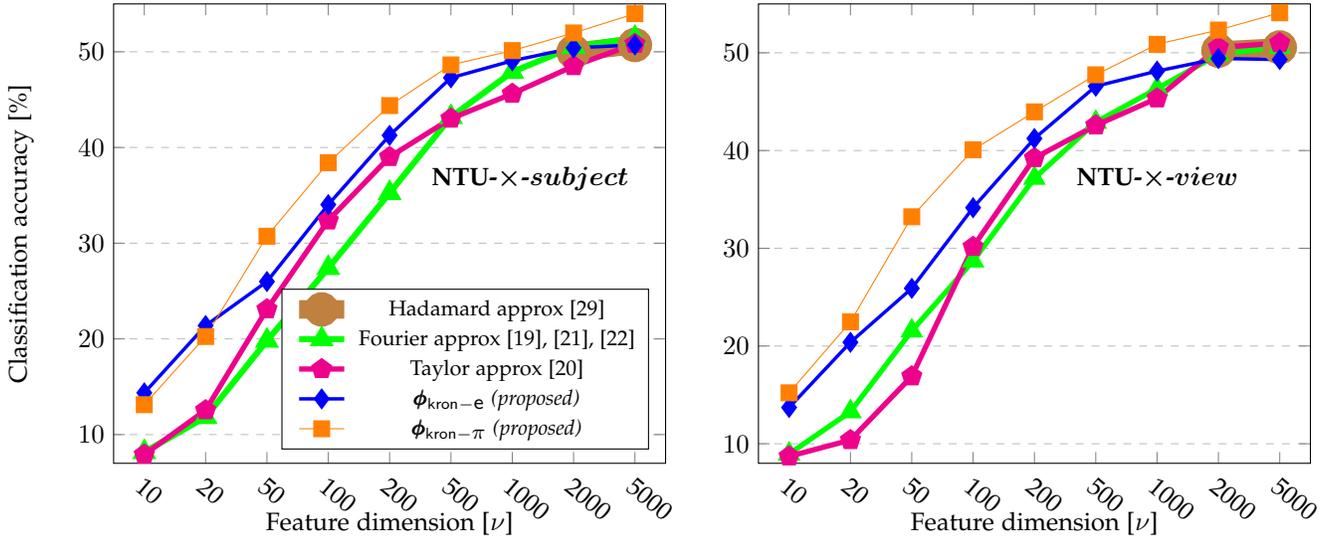
\begin{figure*}
			\vspace{.3 cm}
	\centering
	\tikzset{every mark/.append style={scale=1.5}}
	\tikzset{every axis title/.style={below right,at={(.56,.63)}}}
	\hspace{-9 pt}
	\begin{tikzpicture}
	\begin{axis}[
	width=.49\columnwidth,
	title={\bf NTU-$\boldsymbol{\times}$-$\boldsymbol{subject}$},
	xlabel={Feature dimension [$\nu$]},
	ylabel={Classification accuracy [\%]},
	xmin=0.5, xmax=9.5,
	ymin =  7, ymax = 55,
	ytick={10,20,30,40,50},
	xtick={1,2,3,4,5,6,7,8,9},
	xticklabels = {10, 20, 50, 100, 200, 500, 1000, 2000, 5000},
	xticklabel style={rotate=-40,anchor=north west},
	ymajorgrids=true,
	legend pos=south east,
	grid style=dashed,
	]
	
	\addplot[color=brown,mark=*,line width = 7 pt]
	coordinates{(8,49.98)(9,50.71)};\addlegendentry{Hadamard \cite{Fastfood}}
	
	\addplot[color=green,mark=triangle*, line width = 2.3 pt]
	coordinates{(1,8.17)(2,11.83)(3,19.81)(4,27.41)(5,35.24)(6,43.18)(7,47.9)(8,50.51)(9,51.46)};\addlegendentry{Fourier \cite{RR:NIPS07,Vedaldi:BMVC,Vedaldi_tPAMI}}
	
	\addplot[color=magenta,mark=pentagon*, line width = 2 pt]
	coordinates{(1,7.91)(2,12.55)(3,23.1)(4,32.38)(5,39.01)(6,43.02)(7,45.62)(8,48.51)(9,50.83)};\addlegendentry{Taylor \cite{KK:AISTATS13}}
	
	\addplot[color=blue,mark=diamond*,line width = 1.3]
	coordinates{(1,14.36)(2,21.39)(3,25.99)(4,34.03)(5,41.27)(6,47.28)(7,49.06)(8,50.41)(9,50.72)};\addlegendentry{$\pphi_{\sf{kron}-{\displaystyle e}}$ \emph{(proposed)} \cite{ker_approx}}
	
	\addplot[color=orange,mark=square*]
	coordinates{(1,13.12)(2,20.25)(3,30.73)(4,38.41)(5,44.38)(6,48.63)(7,50.12)(8,51.97)(9,53.98)};\addlegendentry{$\pphi_{\sf{kron}-{\displaystyle \pi}}$ \emph{(proposed)}}
	
	\legend{{\footnotesize Hadamard approx \cite{Fastfood}},{\footnotesize Fourier approx \cite{RR:NIPS07,Vedaldi:BMVC,Vedaldi_tPAMI}},{\footnotesize Taylor approx \cite{KK:AISTATS13}},{\footnotesize $\pphi_{\sf{kron}-{\displaystyle e}}$ \emph{(proposed)}},{\footnotesize $\pphi_{\sf{kron}-{\displaystyle \pi}}$ \emph{(proposed)}}} 
	
	\end{axis}
	\end{tikzpicture}\hspace{8 pt}
	\begin{tikzpicture}
	\begin{axis}[
	width=.49\columnwidth,
	title={\bf NTU-$\boldsymbol{\times}$-$\boldsymbol{view}$},
	xlabel={Feature dimension [$\nu$]},
	xmin=0.5, xmax=9.5,
	ymin =  8, ymax = 55,
	ytick={10,20,30,40,50},
	xtick={1,2,3,4,5,6,7,8,9},
	xticklabels = {10, 20, 50, 100, 200, 500, 1000, 2000, 5000},
	xticklabel style={rotate=-40,anchor=north west},
	ymajorgrids=true,
	grid style=dashed,
	]
	
	\addplot[color=brown,mark=*,line width = 7 pt]
	coordinates{(8,50.22)(9,50.55)};
	
	\addplot[color=green,mark=triangle*,line width = 2 pt]
	coordinates{(1,8.96)(2,13.28)(3,21.55)(4,28.71)(5,37.17)(6,42.9)(7,46.34)(8,49.84)(9,50.6)};

	\addplot[color=magenta,mark=pentagon*, line width = 2 pt]
	coordinates{(1,8.71)(2,10.38)(3,16.9)(4,30.15)(5,39.23)(6,42.58)(7,45.34)(8,50.54)(9,51)};
	
	\addplot[color=blue,mark=diamond*,line width = 1.3 pt]
	coordinates{(1,13.71)(2,20.38)(3,25.9)(4,34.15)(5,41.23)(6,46.58)(7,48.13)(8,49.42)(9,49.31)};
	\addplot[color=orange,mark=square*]
	coordinates{(1,15.21)(2,22.47)(3,33.22)(4,40.08)(5,43.96)(6,47.75)(7,50.87)(8,52.33)(9,54.09)};
	\end{axis}
	\end{tikzpicture}

	\caption{\normalsize Big data regime ($\sim10^4$ samples). Experiments on 
	the NTU RGB+D dataset \cite{Shahroudy:CVPR16} adopting either the 
	cross-subject or the cross-view protocol. In each case, we monitor the changes in 
	action recognition accuracy as a function of the feature dimensionality $\nu$. 
	Across figures, the same color refers to same method. \textit{Note that, 
	differently from Figures \ref{fig:small} and \ref{fig:medium}, the size of the 
	dataset does not allow to directly train a kernel machine, and approximated 
	schemes are obliged.} Best viewed in color.}\label{fig:big}
\end{figure*}


If we compare all the methods analyzed in Figures \ref{fig:small}, \ref{fig:medium} and \ref{fig:big}, we can find a common behavior, that is a growth of accuracy while $\nu$ increases. This it theoretically reasonable because $\pphiP$ and $\pphiE$, as well as the alternative methods \cite{RR:NIPS07,Vedaldi:BMVC,Vedaldi_tPAMI,Fastfood,KK:AISTATS13,Ring:PRL16,ker_approx} are guaranteed to provide a better approximation for a bigger $\nu$. In our case, the reason is that the bound on the variance is $O(1/\nu^3)$. 

As another piece of evidence for correctness of the approximations, we can notice that with a feature dimensionality $\nu \geq 1000$, the performance of each single method is close to the remaining ones and, globally, they are able to mimic the classification accuracy of an exact kernel machine trained in either the small (UTKinect, Florence3D, MSR-$pairs$, MSR-Action3D, G3D) or medium (HDM-05 and MSRC-Kinect12) data regime. The previous claim is also corroborated from the fact that, in those datasets, when $\nu > 500,1000$, we observe a plateau of accuracies since all methods tend to approach the horizontal asymptote given by the exact kernel method (black dotted line). Differently, while moving to the bigger NTU RGB+D, the previous plateau disappears, meaning that an additional increase in the feature dimensionality could be beneficial for improving action recognition.

However, we can observe an interesting pattern which is, in general, common to all datasets for the case $\nu < 200$: at low feature dimensionality (such as 10 or 20), the proposed approximations $\pphiP$ and $\pphiE$ are remarkably superior in performance with respect to all other competitors which are outperformed by margin. For instance, +10\% on Florence3D for $\nu = 10$, +14\% on MSR-Action3D for $\nu = 20$, +9\% on G3D when $\nu = 50$ and more than +10\% on HDM-05$_{all}$ when $\nu = 100$. Moreover, if comparing $\pphiP$ and $\pphiE$, we can observe that, in the small data regime (Figure \ref{fig:small}) the two methods are more or less equivalent. 
Instead, in the middle and big data regime (Figures \ref{fig:medium} and \ref{fig:big}), $\pphiE$ is systematically outperformed by $\pphiP$. 

Finally, as anticipated, an interesting collateral result of our work consists in the possibility to compare the previously proposed methods \cite{RR:NIPS07,Vedaldi:BMVC,Vedaldi_tPAMI,Fastfood,KK:AISTATS13} within a common benchmark. Indeed, despite \cite{Fastfood} shows a solid performance which is always able to match the exact kernel machine and all the other competitors, such approach is limited by the impossibility to obtain a low-dimensional feature representation. Differently, the Fourier \cite{RR:NIPS07,Vedaldi:BMVC,Vedaldi_tPAMI} and Taylor-based methods \cite{KK:AISTATS13} show an oscillating performance where, frequently, one outperforms the other, even by margin. In this respect, the solidity of our methods, which is always top scoring, can be concretely appreciated as an advantage.

\section{Benchmarking the state-of-the-art in 3D action recognition}\label{sez:soa}

In this Section we will compare our proposed approximating schemes $\kea$ and $\kpa$ not only with previously proposed approximations \cite{RR:NIPS07,Vedaldi:BMVC,Vedaldi_tPAMI,Fastfood,KK:AISTATS13}, but also against state-of-the-art approaches for 3D action recognition from skeletal data. Additionally, we will provide the results obtained through our proposed perceptron heuristics $\boldsymbol{\phi}_{\rm P}$.


Before presenting the results, we will briefly discuss the methods involved in the comparison, both kernel methods and feature learning-based approaches.

{\bf Kernel methods.}  We compare against the Fisher vectors-based encoding of \cite{quads} and the Lie group representation \cite{Vemulapalli:CVPR14} and related Lie algebra embedding \cite{Vemulapalli:CVPR16} of rototranslations. We also compare against the combination of multiple non-linear RBF kernels (Ker-RP-RBF) \cite{Wang:ICCV15}, the sequence and dynamics compatibility kernels (SCK + DCK) \cite{ECCV16} and Hankel matrices combined with either HMM (H-HMM) \cite{Camps:ACCV14} or geodesic nearest neighbors method with class-protypes (H-prototypes) \cite{Camps:CVPR16}. Also, we consider the nearest neighbor classification performed in \cite{Weng:CVPR17} through a spatio-temporal Bayesian kernel similarity. Since our approach is covariance-based, we benchmark the temporal pyramid of covariance descriptors ($t$-COV-pyramid) of \cite{egizi}, Bregman-divergence \cite{Harandi:CVPR14} and the kernelized covariance operator (Ker-COV) \cite{Cavazza:ICPR16}. Despite \cite{Minh:CVPR16} applies a similar approximated-covariance paradigm, the published results only pertain to image classification. For completeness, we run the original code and applied it to 3D action recognition, denoting with rnd-logHS and QMC-AlogHS the approaches which exploit either random sampling or Quasi-Monte Carlo integration.  

{\bf Feature learning approaches.} We compete against the following recurrent architectures: the RNN fed on the raw joints data (J-RNN) \cite{Shahroudy:CVPR16} with its body part-aware variant \cite{Du:CVPR15} and we consider Long-Short Term Memory units fed by either raw joints (J-LSTM) \cite{Shahroudy:CVPR16} and its improvements J-LSTM-$a$ \cite{Liu:ECCV16} and J-LSTM$^2$-$a$ \cite{Liu:CVPR17}, which adopt either a shallow or a deep attention module, respectively. We compare against the ensemble of deep models given by RNN-tree \cite{RNNtree:ICCV17} and TSLSTM \cite{TSLSTM:ICCV17} \\
We consider the architectures proposed in \cite{SPDnet} and \cite{deeplie} which embed a structured input data matrix within a deep net: \cite{SPDnet} trains a deep neural network on top of covariance matrices (SPD-Net) and \cite{deeplie} trains on top of rotation matrices. We also compete against LieNet-3B, the 3 blocks configuration that is superior to other investigated in \cite{deeplie}.\\
Also, we benchmark our approach against a few other methods which computes dynamic images (DI), image-like data structures from the joint data to encode the kinematics, and exploit them to train a convolutional neural network. Namely, we consider the J-DI$_E$-CNN \cite{JCNN1} that exploits the Euclidean distance function between joints, J-DI$_\theta$-CNN \cite{JCNN2} that extract DI from rototranslational representations and J-DI$_v$-CNN \cite{Ke:CVPR17} that does the same from velocities, approximated with finite differences.


At the same time, we report the best performance obtained from Figures \ref{fig:small}, \ref{fig:medium} and \ref{fig:big} related to the Hadamard-  \cite{Fastfood}, Fourier- \cite{RR:NIPS07,Vedaldi:BMVC,Vedaldi_tPAMI} and Taylor-based approximations \cite{KK:AISTATS13}, that we indicate with H-approx, F-approx and T-approx, respectively. Ancillary, we also compare with our proposed approximated feature maps $\pphiE$ and $\pphiP$.


The results are reported in Table \ref{tab:soa}, except for the comparison $\pphi_{\rm P}$ versus \cite{ECCV16} which is presented in Table \ref{tab:ECCV} due to the different experimental protocol adopted from \cite{ECCV16}.


\begin{table*}[h!]
			\vspace{.3 cm}
	\centering

\begin{tabular}{rccc}
\hline
& {\scriptsize Florence3D} & {\scriptsize MSR-$pairs$} & {\scriptsize G3D} \\ \hline\hline
H-approx \cite{Fastfood} & 85.5 & 72.8 & 83.8 \\
F-approx \cite{RR:NIPS07}\cite{Vedaldi:BMVC}\cite{Vedaldi_tPAMI} & 83.4 & 72.2 & 83.4 \\
T-approx \cite{KK:AISTATS13} & 84.2 & 73.6 & \textbf{84.6} \\ \hline
$\pphiE$ \textit{(proposed)} & \textit{\textbf{84.9}} & \textit{73.1} & \textit{83.9} \\
$\pphiP$ \textit{(proposed)} & \textit{84.3} & \textit{\textbf{73.7}} & \textit{83.8} \\ \hline \hline
rnd-LogHS \cite{Minh:CVPR16} & 88.1 & 79.4 & 87.8 \\
QMC-logHS \cite{Minh:CVPR16} & 88.5 & 79.5 & 89.5 \\
J-diff-DI-CNN \cite{Ke:CVPR17} & -- & 90.3 & -- \\
LieNet-3B \cite{deeplie} & -- & -- & 89.1 \\
Lie Group \cite{Vemulapalli:CVPR14} & 90.7 & 91.4 & 91.1 \\
Lie Algebra \cite{Vemulapalli:CVPR16} & \underline{{\bf 91.4}} & 94.7 & 90.9 \\ \hline
\textit{$\pphi_{\rm P}$ (proposed)} & \textit{91.2} &  \underline{\textbf{\textit{95.5}}} & \underline{\textbf{\textit{93.0}}} \\ \hline
\end{tabular}\qquad 
\begin{tabular}{rcc}
\hline
& {\scriptsize MSRC-Kinect12} & {\scriptsize HDM-05$_{\rm all}$} \\ \hline\hline
H-approx \cite{Fastfood} & 92.4 & 63.7 \\
F-approx \cite{RR:NIPS07}\cite{Vedaldi:BMVC}\cite{Vedaldi_tPAMI} & 92.2 & 64.0 \\
T-approx \cite{KK:AISTATS13} & 92.8 & 62.0 \\ \hline
$\pphiE$ \textit{(proposed)} & \textit{92.3} & \textit{65.0} \\
$\pphiP$ \textit{(proposed)} & \textit{\textbf{95.6}} & \textit{\textbf{ 66.5}} \\ \hline \hline
$t$-COV-pyramid \cite{egizi} & 89.2 & -- \\
Bregman-div \cite{Harandi:CVPR14} & 89.9 & 58.2  \\
Ker-RP-RBF \cite{Wang:ICCV15} & 92.3 & 66.2 \\
J-DI$_E$-CNN \cite{JCNN1} & 93.1 & $-$ \\
Ker-COV \cite{Cavazza:ICPR16} & 95.0 & -- \\
rnd-logHS \cite{Minh:CVPR16} & 97.1 & 58.1 \\
QMC-logHS \cite{Minh:CVPR16} & 96.2 & 60.2 \\
SPD-net \cite{SPDnet} & -- & 61.4 \\ \hline
\textit{$\pphi_{\rm P}$ (proposed)} & \underline{\textbf{\textit{98.5}}} & \underline{\textbf{\textit{72.0}}} \\ \hline
\end{tabular}\\\vspace{10 pt}
\begin{tabular}{rcc}
\hline 
& {\scriptsize NTU-$\times$-$subject$} & {\scriptsize NTU-$\times$-$view$} \\\hline\hline
H-approx \cite{Fastfood} & 51.5 & 50.6 \\
F-approx \cite{RR:NIPS07}\cite{Vedaldi:BMVC}\cite{Vedaldi_tPAMI} & 50.7 & 50.6 \\
T-approx \cite{KK:AISTATS13} & 50.8 & 51.0 \\ \hline
$\pphiE$ \textit{(proposed)} & \textit{50.7} & \textit{49.4} \\
$\pphiP$ \textit{(proposed)} & \textbf{\textit{54.0}} & \textbf{\textit{54.1}} \\ \hline \hline
Fisher Vectors \cite{quads} & 38.6 & 41.4 \\
Lie Group \cite{Vemulapalli:CVPR14} & 50.1 & 52.8 \\  
J-RNN \cite{Shahroudy:CVPR16}  & 56.3 & 64.0  \\
J-RNN-parts \cite{Du:CVPR15}  &  59.1 &  64.1 \\
LieNet-3B \cite{deeplie} & 61.4 & 67.0 \\
J-LSTM \cite{Shahroudy:CVPR16} & 60.7 & 67.3  \\
J-LSTM-$a$ \cite{Liu:ECCV16} & 69.2 & 77.7 \\
J-DI$_E$-CNN \cite{JCNN1} & 73.4 & 75.2 \\
J-LSTM$^2$-$a$ \cite{Liu:CVPR17} & 74.4 & 82.8 \\
TS-LSTM \cite{TSLSTM:ICCV17} & 74.6 & 81.3 \\
RNN-tree \cite{RNNtree:ICCV17} & 74.6 & 83.2 \\
J-DI$_\theta$-CNN \cite{JCNN2} & 76.2 & 82.3 \\
J-DI$_v$-CNN \cite{Ke:CVPR17} & {\bf 79.6} & {\bf 84.8} \\ \hline
\textit{$\pphi_{\rm P}$ (proposed)} & \textit{60.9} & \textit{63.4} \\ \hline
\end{tabular}\qquad
\begin{tabular}{rccc}
\hline
& {\scriptsize MSR-Action3D} & {\scriptsize  HDM-05$_{14}$} &  {\scriptsize UTKinect} \\ \hline\hline
H-approx \cite{Fastfood} & 88.4 & 89.2 & 83.9 \\
F-approx \cite{RR:NIPS07}\cite{Vedaldi:BMVC}\cite{Vedaldi_tPAMI} & 88.2 & 88.6  & 84.0 \\
T-approx \cite{KK:AISTATS13} & 89.6 & 88.9 & 84.0 \\ \hline
$\pphiE$ \textit{(proposed)} & \textit{89.5} & \textit{89.6} & \textit{\textbf{84.4}} \\
$\pphiP$ \textit{(proposed)} & \textit{\textbf{89.9}} & \textit{\textbf{89.9}} & \textit{84.0} \\ \hline \hline
$t$-COV-pyramid \cite{egizi} & 74.0 & 91.5 & -- \\
H-HMM \cite{Camps:ACCV14} & 89.0 & -- & 86.8 \\
rnd-logHS \cite{Minh:CVPR16} & 91.5 & 88.5 & 89.7 \\
QMC-logHS \cite{Minh:CVPR16} & 90.6 & 85.4 & 91.3 \\
H-prototypes \cite{Camps:CVPR16} & 94.7 & 86.3 & \underline{{\bf 100}} \\
TS-LSTM \cite{TSLSTM:ICCV17} & -- & -- & 97.0 \\
J-LSTM \cite{Liu:ECCV16} & 94.8 & -- & 97.0 \\
Ker-RP-RBF \cite{Wang:ICCV15} & 96.9 & 96.8 & -- \\
ST-BNN \cite{Weng:CVPR17} & 94.8 & -- & 98.0 \\
Ker-COV \cite{Cavazza:ICPR16} & 96.8 & 98.1 & --\\ \hline		
\textit{$\pphi_{\rm P}$ (proposed)} & \underline{\textbf{\textit{97.4}}} & \underline{\textbf{\textit{99.1}}} & \textit{98.3} \\ \hline
\end{tabular}  
    
	\caption{\normalsize Classification accuracies [$\%$] for 3D action recognition. 
	For each table, the top part present the performance achieved by $\pphiP$ and 
	$\pphiE$ against other alternative approximating schemes 
	\cite{RR:NIPS07,Vedaldi:BMVC,Vedaldi_tPAMI,Fastfood,KK:AISTATS13}: within 
	this class of methods, the best accuracy is highlighted in bold. At the same time, 
	in the bottom part of each table, $\pphi_{\rm P}$ is compared against 
	state-of-the-art approaches and, among them, the best performance is marked 
	by bold and underlined. All the performance achieved by methods proposed in 
	this paper ($\pphiP,\pphiE$ and $\pphi_{\rm P}$) are in italic.}\label{tab:soa}

\end{table*}

\begin{table}[h!]
			\vspace{.3 cm}
	\begin{tabular}{rcccc}
		\hline
		& {\tiny Florence3D$^\star$}              & {\tiny UTKinect}        & {\tiny MSR-Action3D$^\star$}            & {\tiny MSR-Action3D} \\ \hline\hline
		SCK \cite{ECCV16}          & 92.98                   & 96.1            & 90.72                   & 93.5                   \\
		DCK \cite{ECCV16}         & 93.03                   & 97.5            & 86.30                   & 91.7                   \\
		SCK+DCK \cite{ECCV16}   & 95.23                   & 98.2 & 91.45                   & 94.0                   \\\hline
		\emph{$\pphi_{\rm P}$ (proposed)} & \underline{\textit{\textbf{97.25}}} & \underline{\textit{\textbf{98.3}}} & \underline{\textit{\textbf{96.30}}} & \underline{\textit{\textbf{97.4}}} \\\hline 
	\end{tabular}
	\caption{\normalsize Classification accuracies [$\%$] of $\pphi_{\rm P}$ 
	against \cite{ECCV16}. Best results are bold and underlined, the symbol 
	$^\star$ indicates that we used the alternative training/testing split adopted in 
	\cite{ECCV16}.}\label{tab:ECCV}
\end{table}

\subsection{Discussion}

While inspecting the performance of the approximated feature maps $\pphiE$ and $\pphiP$, in Table \ref{tab:soa}, we can appreciate a little improvement ($< 1\%$) over the alternative methods \cite{RR:NIPS07,Vedaldi:BMVC,Vedaldi_tPAMI,Fastfood,KK:AISTATS13} in the small data regime of Florence3D, MSR-$pairs$, MSR-Action3D, HDM-05$_14$ and UTKinect (for G3D T-approx is about 1\% better). This is coherent with what we found out in Section \ref{sez:against}: since the aforementioned performance is mainly all achieved by $\nu = 5000$, in such a case, all methods seem to converge towards the performance of an exact kernel machine fed by \eqref{eq:K}. Differently, in the remaining datasets, either in the medium or big data regime, we register a significant boost in performance of $\pphiE$ and $\pphiP$ over the other approaches.

Still, we can spot how, in certain cases, $\pphiE$ and $\pphiP$ are better than methods which have been explicitly designed for action recognition: remember that, in theory, those approximations hold for any type of $d \times d$ data input. For instance, on MSR-Action3D, $\pphiE$ and $\pphiP$ improves \cite{egizi} by about +15\% and, on the NTU RGB+D dataset with the cross-subject protocol, the performance of \cite{Vemulapalli:CVPR14} and \cite{quads} is improved by +4\% and +16\%.  Eventually, on the NTU-$\times$-$subject$, the performance scored by  $\pphiE$ and $\pphiP$ is almost on par with respect to the deep recurrent neural networks J-RNN and J-RNN-parts. Furthermore, in the middle data regime of MSRC-Kinect12 and HDM-05$_{all}$,  $\pphiE$ and $\pphiP$ (and, in general, all the other approximated feature maps hereby considered) are  scoring  better than \cite{egizi,Harandi:CVPR14,Wang:ICCV15,JCNN1} on  MSRC-Kinect12. For what concerns HDM$_{all}$,  $\pphiP$ is even able to beat by 5\% the SoA deep learning method SPD-net \cite{SPDnet}. Such trend can be motivated by the fact that, in the middle data regime ($\sim 10^4$), the data instances are sufficiently rich to train satisfactory decision boundaries in a max margin sense, while not enough to effectively train deep models which suffer from their over-parametrization.

While moving from either $\pphiE$ or $\pphiP$ to $\pphi_{\rm P}$, we \emph{always} observe a growth in performance, the latter being about +2\% in the worst case and about +22\% in the best one. Precisely, in the small data regime, we improved previously published state-of-the-art classification results by +0.5\% on MSR-Action3D, +0.8\% on MSR-$pairs$, +1\% on HDM-05$_{14}$ and by +2.1\% on G3D. 

At the same time, the gap in accuracy between $\pphi_{\rm P}$ and $\pphiE,\pphiP$ grows as the size of the dataset increases: such correlation is clearly a matter of the well known fact that feature learning benefits from more data. Again, the middle data regime seems the ideal operative setting for $\pphi_{\rm P}$, since, to the best of our knowledge, the previously published state-of-the-art performance on MSRC-Kinect12 by +2.3\% (with respect to Ker-RP-RBF \cite{Wang:ICCV15}) and by +10.6\% on HDM-05$_{all}$.

On the NTU RGB+D experiments, $\pphi_{\rm P}$ improves (by margin) Fisher vectors \cite{quads}, Lie group representation \cite{Vemulapalli:CVPR14} as well as the deep J-RNN and J-RNN-parts on the NTU-$\times$-$subjects$. However, when comparing with the performance of LSTM- and CNN-based methods, $\pphi_{\rm P}$ shows a suboptimal performance. This trend can be justified in two ways.\\
On the one hand, we are applying a shallow architecture with just one hidden layer while, for instance, J-DI$_v$-CNN and JCNN2 uses multiple deep convnets in parallel and J-LSTM$^2$-$a$ conditions a deep LSTM on the output of another deep LSTM network.\\
On the other hand, all LSTM-based methods and J-DI$_v$-CNN access all the raw coordinates for each given timestamps: therefore, since we train the architecture of \cite{kermeetfeat} on covariance matrices, we can say that we are using much less data that are reduced by a factor of approximatively $1/100$, being $100$ the typical temporal length for the sequences on the NTU RGB+D dataset.

Despite the previous two points are a drawback in terms of classification accuracies, they results in the following operative advantages. \\
First, since the  architecture of \cite{kermeetfeat} is shallow, there is no need for GPU acceleration neither for inference (which is nevertheless real-time), nor for the training stage (which, even on CPU, only lasts less than one hour, as opposed to one day, for instance, for the LSTM networks to be trained \cite{Liu:CVPR17}). Therefore, our system achieves a clear portability for deployment in real-world applications that requires real-time and scalable recognition capabilities.\\
Second, our representation is very compact: the experiments reported in Table \ref{tab:ECCV}, we are able to always overcome SCK and DCK in performance, even using a feature representation which is about 100 times more compact. Even on the NTU RGB+D dataset, we train the coefficients of the support vectors on top of the hidden representation of \cite{kermeetfeat} where its size is fixed to $2^8$. Having only two sets of weighted elements is a very favorable operative condition as opposed to stacking several convolutional layers \cite{JCNN1,JCNN2,Ke:CVPR17} or allocating high-dimensional tensors for back-propagating through times and train the architectures of \cite{Du:CVPR15,Shahroudy:CVPR16,Liu:ECCV16,Liu:CVPR17}. 

This certifies in empirical terms the benefits of learning instead of sampling weights since although being a simple heuristics, the improvements in performance justifies the soundness of our proposed $\pphi_{\rm P}$.

\section{Conclusions \& future work}\label{sez:conc}

This paper presented $\pphiP$ a novel approximation scheme for the RBF kernel function, which was shown to be superior to other approximations \cite{RR:NIPS07,Vedaldi:BMVC,Vedaldi_tPAMI,Fastfood,KK:AISTATS13} in terms of better variance bound and classification accuracy, being the computational cost almost equal. 

In a broad experimental evaluation among state-of-the-art competitors over publicly available action recognition datasets, our method generally assesses its superiority in terms of classification accuracy. Such favorable performance is also obtained by means of a very compact model, that is characterized by a simple \& fast training procedure, especially if compared to deep learning methods.

We opted for a trade-off ($\rho$ to be Geometric distributed of parameter $\theta = .9$) which slightly penalizes variance in favor of computational efficiency, but with practically no impact on classification performance.

As future work, we aim at investigating a couple of topics. In theoretical terms, We will try to devise an improved bound on the variance, while, from  the application standpoint, we want to apply our approximation to other computer vision tasks, such as object categorization.

\ifCLASSOPTIONcaptionsoff
  \newpage
\fi
%
%


\bibliographystyle{IEEEtran}
\bibliography{fonti}
%
%
%
%
%
%

\newpage

\begin{IEEEbiography}[{\includegraphics[width=1in,height=1.25in,clip]{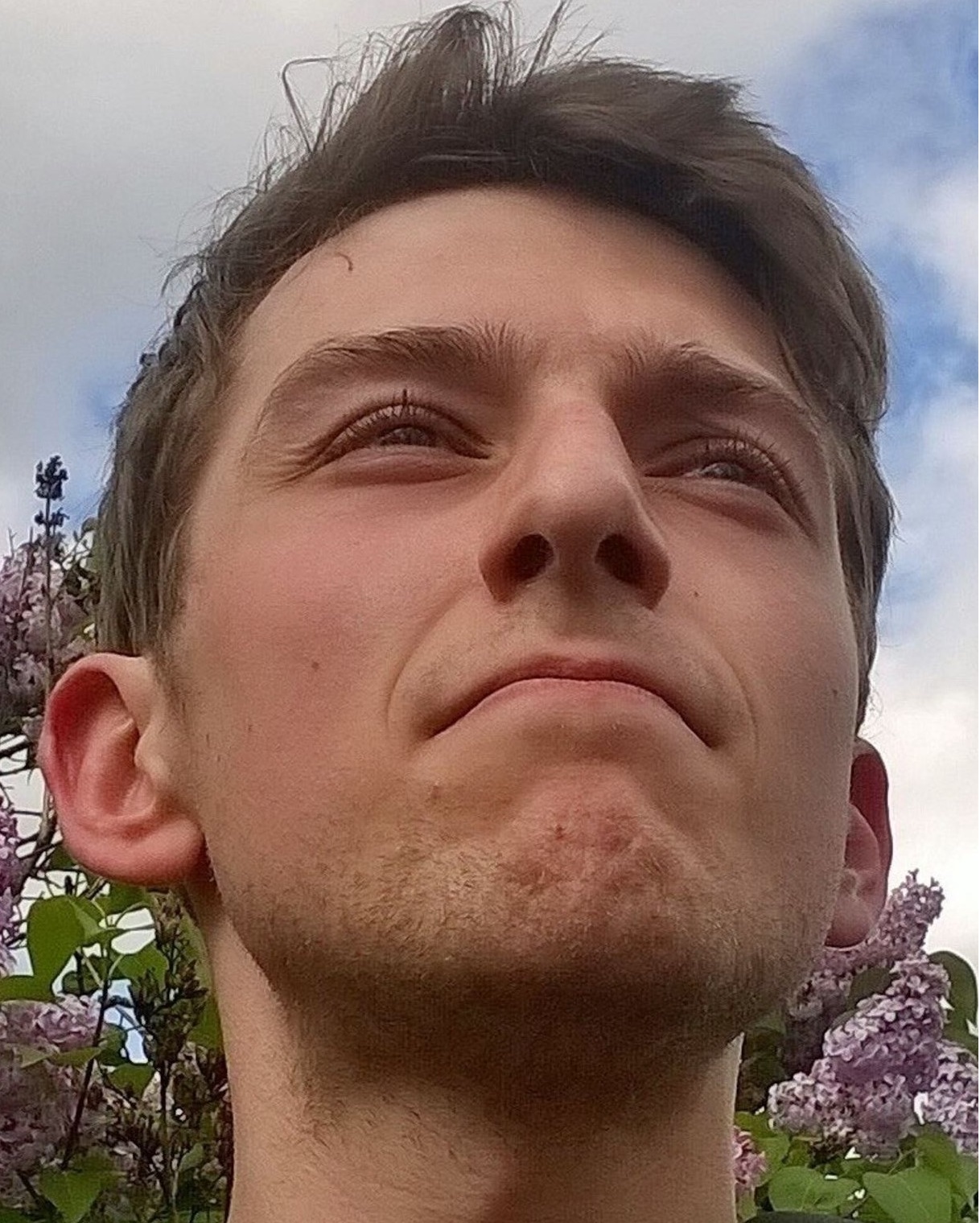}}]{Jacopo
 Cavazza} received his M.Sc. degree in Mathematics from the University of Genova, 
Italy, in 2014 (magna cum laude). He is currently a Fellow PhD of the Department of Naval, Electric, 
Electronic and Telecommunication Engineering (DITEN) of the University of Genova, 
operating at the Pattern Analysis and Computer Vision Department (PAVIS), Istituto 
Italiano di Tecnologia (IIT), Italy, Genova under the supervision of Prof. Vittorio 
Murino. His research interests deals with machine learning and computer vision, 
particularly, video-surveillance and human action analysis and intention prediction. 
\end{IEEEbiography}%
\begin{IEEEbiography}[{\includegraphics[width=1in,height=1.25in,clip]{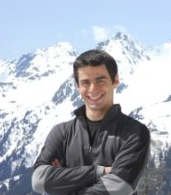}}]{Pietro Morerio} Pietro Morerio received his M. Sc. in Theoretical Physics from the University of Milan (Italy) in 2010 (summa cum laude). He was Research Fellow at the University of Genoa (Italy) from 2011 to 2012, working in Video Analysis for Interactive Cognitive Environments. He pursued a PhD degree in Computational Intelligence at the same institution in 2016. He is currently a Postdoctoral Researcher at Istituto Italiano di Tecnologia (IIT). His research focuses on machine learning and computer vision.
\end{IEEEbiography}%
\begin{IEEEbiography}[{\includegraphics[width=1in,height=1.25in,clip]{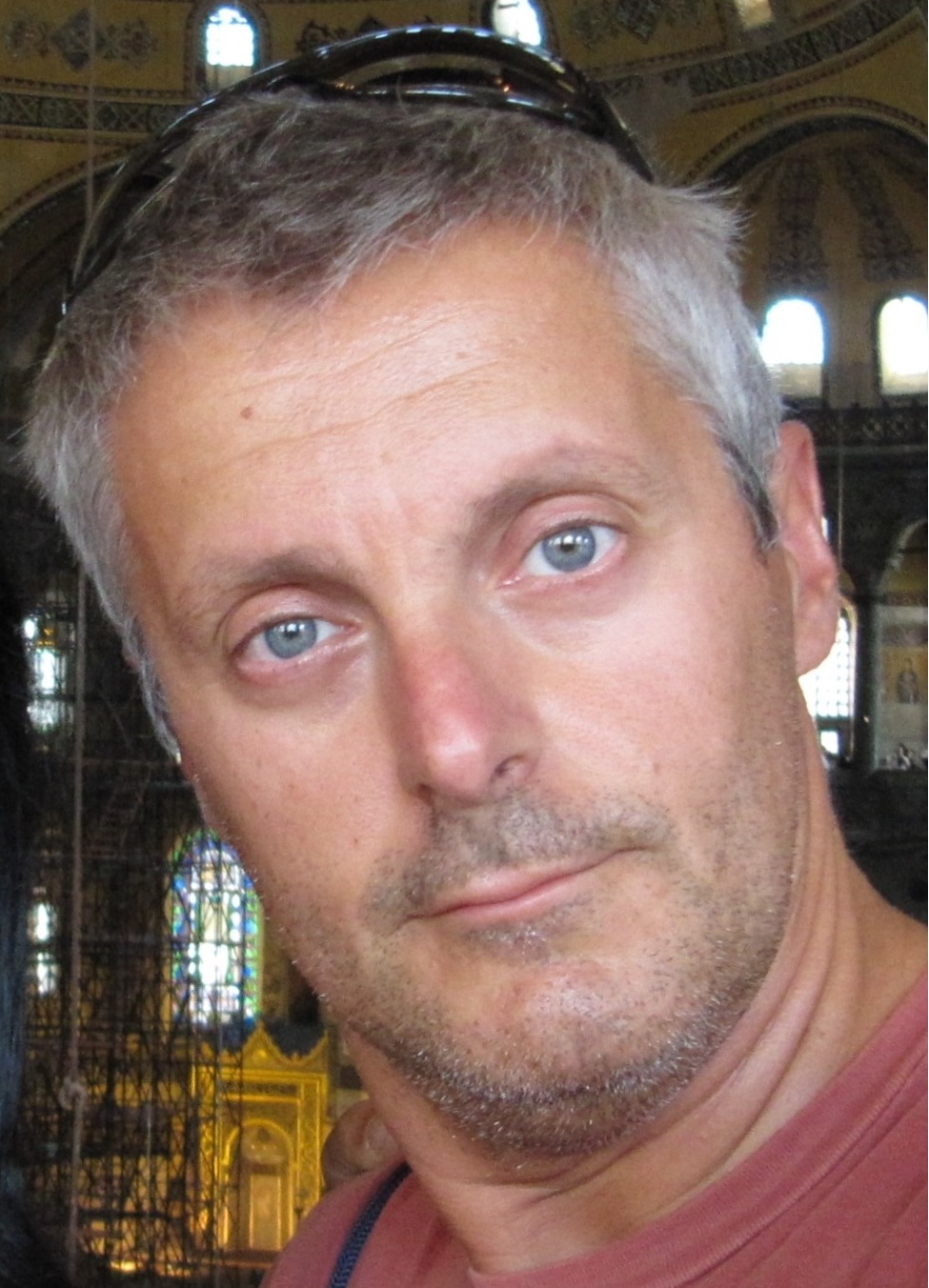}}]{Vittorio
 Murino} (SM'02) received the Laurea degree in electronic engineering 
	and the Ph.D. degree in electronic engineering and computer science from the 
	University of Genova, Italy, in 1989 and 1993, respectively.
	He is a full professor at the University of Verona, Italy since 2000. Since 1998, he 
	has been with the University of Verona where he held the Chair of the 
	Department of Computer Science of this University from 2001 to 2007. 
	He is currently with the Istituto Italiano di Tecnologia, as director of Pattern 
	Analysis and Computer Vision (PAVIS) department, involved in computer vision, 
	machine learning, and image analysis activities. He has co-authored over 400 
	papers published in refereed journals and international conferences. His current 
	research interests include computer vision, pattern recognition, and machine 
	learning, more specifically, statistical and probabilistic techniques for image and 
	video processing, with applications on (human) 	behavior analysis and related 
	applications such as video surveillance, biomedical imaging, and bioinformatics.
	He is also an Associate Editor of \textit{Computer Vision and Image Understanding}, \textit{Machine Vision \& Applications}, and \textit{Pattern Analysis and Applications} journals. He is also an IAPR Fellow since 2006.
\end{IEEEbiography}

\vfill


%
%
%



\end{document}